%% file: main.tex
\documentclass[11pt]{article}
\usepackage{preprint}

\input{math_commands.tex}

\usepackage{hyperref}
\usepackage{url}

\usepackage{amsmath, amssymb, amsthm}
\usepackage{hyperref}
\usepackage{url}
\usepackage{booktabs}
\usepackage{graphicx}
\usepackage{subcaption}
\usepackage{enumitem}
\usepackage{wrapfig}
\usepackage{multirow}
\usepackage{titletoc}
\usepackage{xcolor}
\usepackage{colortbl}
\usepackage{amssymb}
\usepackage{tabularx}
\usepackage{arydshln}
\usepackage{amsmath}
\usepackage{algorithm}
\usepackage[noEnd=false,indLines=false]{algpseudocodex}

\algrenewcommand{\algorithmicrequire}{\textbf{Input:}}

\algrenewcommand{\algorithmicrequire}{\textbf{Input:}}

\definecolor{blue}{HTML}{4F75D8}
\definecolor{orange}{HTML}{E88942}
\definecolor{green}{HTML}{4DB565}
\definecolor{red}{HTML}{D85454}
\definecolor{purple}{HTML}{856ED0}
\definecolor{brown}{HTML}{A8794F}
\definecolor{pink}{HTML}{DB6DBB}
\definecolor{gray}{HTML}{8A8A8A}
\definecolor{darkgray}{HTML}{444444}
\definecolor{darkblue}{HTML}{113399}
\definecolor{yellow}{HTML}{D8B943}
\definecolor{bluegreen}{HTML}{0D98BA}

\theoremstyle{plain}

\theoremstyle{definition}

\theoremstyle{remark}

\usepackage{listings}
\usepackage{tcolorbox}
\tcbuselibrary{skins,breakable,listings}

\newcommand{\promptarg}[1]{%
  \textcolor{brown}{\ttfamily\detokenize{<#1>}}%
}
\newtcblisting{promptbox}[2][]{
  enhanced,
  breakable,
  listing only,
  listing engine=listings,
  title={#2},
  width=\linewidth,
  colback=black!0.8,
  colbacktitle=pptint,
  coltitle=black,
  colframe=black!30,
  fonttitle=\scriptsize\bfseries\color{darkgray},
  toptitle=1.5mm,
  bottomtitle=1.5mm,
  boxrule=0.6pt,
  titlerule=0.5pt,
  arc=1mm,
  listing options={
    language={},
    basicstyle={\linespread{1}\fontsize{7.5}{8.65}\selectfont\ttfamily},
    lineskip=0pt,
    columns=fullflexible,
    keepspaces=true,
    breaklines=true,
    breakatwhitespace=true,
    breakautoindent=false,
    showstringspaces=false,
    showspaces=false,
    showtabs=false,
    numbers=none,
    frame=none,
    tabsize=4,
    aboveskip=0pt,
    belowskip=0pt,
    escapeinside={(*@}{@*)},
    moredelim={[is][\color{darkblue}]{<<}{>>}}
  },
  #1
}

\title{Improving LLM Collaboration via Multi-Agent Preference Learning}
\runningtitle{Improving LLM Collaboration via Multi-Agent Preference Learning}

\author{%
Shuo Liu\affil{$\dagger$}, Xinzichen Li, Tianle Chen, Christopher Amato\affil{$\dagger$}}
\affiliation{%
Northeastern University, Boston\\
\textsuperscript{$\dagger$}Correspondence to: \texttt{\{liu.shuo2,c.amato\}@northeastern.edu}}

\abstract{%
Several works have explored \textit{multi-agent reinforcement learning} (MARL) in LLM collaboration. However, constructing reliable rewards is difficult in practice, as complete and accurate metrics are often unavailable and hard to aggregate. Preference learning provides an alternative by learning from comparative human or AI feedback. Yet, its extension to multi-agent systems remains underexplored.
To address this gap, we formulate preference-based multi-agent systems (MAS) from decentralized and centralized collaboration perspectives. 
We also introduce a general \textit{multi-agent preference learning} framework (MAPL) to solve these problems. MAPL allows iterative updates by comparing the current solution with decentralized or centralized solutions generated by various agents. We instantiate MAPL using \textit{MARL from human feedback} (MARLHF) with a learned reward model and \textit{multi-agent direct preference optimization} (MADPO).
Experiments on collaborative writing, coding, tool use, and travel planning show that MAPL can improve collaboration quality and efficiency while approaching the performance of MARL with fixed, well-defined rewards.
Within MAPL, MARLHF generally outperforms MADPO on most tasks but remains sensitive to data coverage, agent and comparator models, and the underlying MARL algorithms.}

\begin{document}

\maketitle

\section{Introduction}

Recent advances in agentic AI have transformed LLMs into autonomous agents that can solve complex tasks by interacting with environments \citep{yao2022react,shinn2023reflexion,wang2023voyager,achiam2023gpt,bai2023qwen,liu2024deepseek,shao2024deepseekmath,comanici2025gemini}.

As LLM agents become more diverse and specialized, many studies have explored multi-agent cooperation to tackle complex tasks that are difficult for a single agent. Multi-agent systems (MAS) can be organized in several ways.
For example, agents can play different roles and communicate to collaborate in a centralized manner \citep{li2023camel,qian2024chatdev,du2023improving}. Such coordination can also be directed by a central manager agent \citep{wu2023autogen,hong2024metagpt}. Alternatively, agents can solve subtasks independently on different devices to achieve higher efficiency \citep{zhang2024proagent,liu2026learning,liu2026llm}. In such decentralized MAS, each agent processes only its local context. This lowers the computational requirements for each device and allows agents to be deployed locally, which can also improve data privacy.
Nevertheless, most general-purpose LLMs are not explicitly trained for cooperation and thus often struggle to coordinate effectively with other agents \citep{park2025maporl,cemri2026multi}.

\textit{Multi-agent reinforcement learning} (MARL) has been extensively studied to improve coordination among agents \citep{zhang2021multi,albrecht2024multi,amato2024introduction}. Building on these advances, recent studies applied MARL to optimize LLM interaction \citep{zhao2025stronger,liao2025marft,zhang2025marti,liu2026learning,liu2026llm,wang2026marti,chen2026improving}. However, many approaches rely on hand-crafted rewards based on domain-specific metrics, which may be unavailable or fail to accurately capture the intended task objective \citep{zhu2024decoding,bui2025mapl}.

Preference learning provides an alternative for optimizing LLM agents by leveraging pairwise human feedback without complex reward engineering \citep{christiano2017deep, ziegler2019fine, rafailov2023direct,meng2024simpo}. 
Conventional preference learning methods mostly rely on fixed offline datasets, whereas some online approaches collect fresh preferences during training to reduce the distribution mismatch between the data and the evolving policy \citep{guo2024direct,xiong2023iterative}. 
Yet, extending these methods to LLM collaboration remains challenging. The vast policy space of MAS makes it difficult to collect preference data to adequately cover diverse cooperation schemes \citep{oliehoek2016concise,pan2022plan,yang2021believe,shao2023counterfactual}. Also, the learned rewards may be inaccurate or high-variance when the training data is insufficient, which could undermine the subsequent policy optimization.

To fill the gap, we study preference-based optimization of LLM collaboration. We formulate decentralized collaboration as a \textit{preference-based decentralized partially observable MDP} (\textbf{Pb-Dec-POMDP}) and centralized collaboration as a \textit{preference-based multi-agent partially observable MDP} (\textbf{Pb-MPOMDP}), and highlight their unique challenges. We introduce \textit{multi-agent preference learning} (\textbf{MAPL}), a general framework that iteratively updates the agents' policies by comparing their joint solutions with those by a comparator model. MAPL considers three paradigms based on whether the agents and comparator operate in a decentralized or centralized manner: \textit{decentralized collaboration with a decentralized comparator} (Dec-Dec), \textit{centralized collaboration with a centralized comparator} (Cen-Cen), and \textit{decentralized collaboration with a centralized comparator} (Dec-Cen). We also instantiate MAPL with two approaches: two-stage \textit{multi-agent reinforcement learning from human feedback} (\textbf{MARLHF}) and a direct approach, \textit{multi-agent direct preference optimization} (\textbf{MADPO}).

Our experiments on collaborative writing, coding, tool-use, and travel planning tasks show that MAPL can achieve performance similar to MARL with well-specified rewards. MAPL effectively improves decentralized and centralized collaboration under Dec-Dec and Cen-Cen, whereas its Dec-Cen variants perform poorly in most domains due to information and realizability gaps between the comparator and agents. MARLHF generally outperforms MADPO across most domains, but can exhibit high variance when preference data is limited, the comparator policy differs substantially from the agents', or actor-critic methods are used for policy optimization.

\section{Preliminaries} \label{sec:preliminaries}

\paragraph{BT model in preference learning}
Preference learning optimizes policies from comparative feedback. The Bradley-Terry (BT) model is a commonly used model to parameterize pairwise preference probabilities \citep{bradley1952rank}. Let $\mathcal{X}$ and $\mathcal{Y}$ denote the prompt and response spaces, and let $R^*:\mathcal{X}\times\mathcal{Y}\to\mathbb{R}$ be an oracle reward. For $x\in\mathcal{X}$ and $y,y'\in\mathcal{Y}$,
\begin{equation}
\Pr(y \succ y' \mid x)
=
\frac{\exp(R^*(x,y))}
{\exp(R^*(x,y))+\exp(R^*(x,y'))}
=
\sigma\!\left(R^*(x,y)-R^*(x,y')\right),
\label{eq:BTmodel}
\end{equation}
where $\sigma(z)=1/(1+\exp(-z))$. In practice, a preference learning method typically uses a dataset $\mathcal{D}={(x_j,y_j^w,y_j^l)}_{j=1}^{m}$, where $y_j^w$ is preferred to $y_j^l$ under prompt $x_j$.

\paragraph{Two-stage RLHF approaches}
RLHF aligns LLMs from human preference \citep{christiano2017deep,ziegler2019fine,instructgpt,bai2022training,touvron2023llama}. Classic RLHF approaches involve two stages \citep{christiano2017deep,ziegler2019fine}, where a reward model $\widetilde{R}_\psi$ is first trained from pairwise preferences, then the policy $\pi_\theta$ is optimized with KL regularization against a reference policy $\pi_{\mathrm{ref}}$,
\begin{equation}
\begin{gathered}
\mathcal{L}_{\mathrm{RM}}(\psi)
=
-\mathbb{E}_{(x,y^w,y^l)\sim\mathcal{D}}
\left[
\log \sigma\left(
\widetilde{R}_\psi(x,y^w)-\widetilde{R}_\psi(x,y^l)
\right)
\right],
\\
\mathcal{L}_{\mathrm{RL}}(\theta)
=
-\mathbb{E}_{x\sim\varrho,\,y\sim\pi_\theta(\cdot\mid x)}
\left[
\widetilde{R}_\psi(x,y)
-
\eta \log
\frac{\pi_\theta(y\mid x)}
{\pi_{\mathrm{ref}}(y\mid x)}
\right].
\end{gathered}
\label{eq:rlhf}
\end{equation}
Here, $\varrho$ denotes the prompt distribution and $\eta>0$ is the KL penalty coefficient. We use RLHF to refer specifically to this two-stage, reward-modeling approach in the following parts.

\paragraph{One-stage DPO approaches}
The two-stage RLHF pipeline requires both reward-model training and RL optimization, which is sensitive to hyperparameters and introduces substantial computational overhead \citep{chaudhari2025rlhf}. \citet{rafailov2023direct} proposed an efficient approach, DPO, that directly optimizes the policy without explicitly fitting a reward model,
\begin{equation}
\mathcal{L}_{\mathrm{DPO}}(\theta)
=
-\mathbb{E}_{(x,y^w,y^l)\sim\mathcal{D}}
\left[
\log \sigma\left(
\eta \log
\frac{\pi_\theta(y^w\mid x)}
{\pi_{\mathrm{ref}}(y^w\mid x)}
-
\eta \log
\frac{\pi_\theta(y^l\mid x)}
{\pi_{\mathrm{ref}}(y^l\mid x)}
\right)
\right].
\label{eq:dpo}
\end{equation}
Under the standard DPO derivation, this objective corresponds to the same KL-regularized preference optimization problem as RLHF, while in practice, their performance can differ, depending on tasks, model realizability, and optimization \citep{shi2025understanding,razin2025your}.

\section{MAS with Preference Feedback} \label{sec:background}

We study cooperative \textit{multi-agent systems} (MAS), where LLM agents collaborate to solve a class of tasks \citep{qian2024chatdev,hong2024metagpt,chen2026improving,liu2026learning,liu2026llm,zhao2025stronger}. While most work assumes scalar rewards, we consider MAS with preference feedback.

\subsection{Preference-based Decentralized MAS}

Decentralized MAS support efficient and scalable execution, as LLM agents can run inference in parallel across different nodes~\citep{zhang2024proagent,liu2026learning,liu2026llm}. In this setting, each task is specified by language prompts and presented to the agents. At each turn, every agent generates a response based on its local context, and the individual responses are aggregated into a solution. Users, tools, and other models in the system then provide new requirements, feedback, or constraints that form the prompts for the next turn. This interaction continues until the task is completed or the turn limit is reached. For each episode, two trajectories are sampled, and an annotator compares their returns, producing preferences for policy optimization.

We formulate this problem as a \textit{preference-based decentralized partially observable Markov decision process} (Pb-Dec-POMDP), denoted as $\langle \mathcal{I},\mathcal{S},\{\mathcal{O}_i\}_{i\in\mathcal{I}},\{\mathcal{A}_i\}_{i\in\mathcal{I}},T,O,R^*,H,b_0,P_{R^*}\rangle$ \citep{oliehoek2016concise}. Here, $\mathcal{I}=\{1,\ldots,n\}$ is a set of $n$ LLM agents, where each agent $i$ follows an individual policy $\pi_i$, and all policies form a joint policy $\boldsymbol{\pi}=(\pi_1,\ldots,\pi_n)$. At each turn $t$, the state $s_t=(s_t^{\mathrm{sys}},s_t^{\mathrm{usr}})\in\mathcal{S}$ consists of an accessible part from the system $s_t^{\mathrm{sys}}\in\mathcal{S}^{\mathrm{sys}}$ and a latent part from users $s_t^{\mathrm{usr}}\in\mathcal{S}^{\mathrm{usr}}$, with initial state distribution $b_0\in\Delta(\mathcal{S})$. The underlying state cannot be directly observed by the agents. Each agent $i$ receives a natural-language prompt as its local observation $o_{i,t}\in\mathcal{O}_i$, and all agents' observations form a joint observation $\mathbf{o}_t=\{o_{1,t},\cdots, o_{n,t}\}\in\mathcal{O}$. The observation $o_{i,t}$ at turn $t$ provides a partial and potentially noisy view of $s_t$. Each agent generates a response as an action $a_{i,t}\sim\pi_i(\cdot|h_{i,t}),a_{i,t}\in\mathcal{A}_i$, which depends on its observation-action history $h_{i,t}=\{o_{i,0},a_{i,0},\ldots,a_{i,t-1},o_{i,t}\}$. The responses of all agents form the joint action $\mathbf{a}_t=\{a_{1,t},\ldots,a_{n,t}\}\in\mathcal{A}$, which induces a transition in the environment $s_{t+1}\sim T(\cdot| s_t,\mathbf{a}_t)$. The observation is then emitted as $\mathbf{o}_{t+1}\sim O(\cdot| s_{t+1},\mathbf{a}_t)$ at the next turn $t+1$, where each agent receives its local observation $o_{i,t+1}$. We use $H$ to denote the finite episode horizon.

In each episode of the Pb-Dec-POMDP, two trajectories $\tau,\tau'\in \Upsilon$ are sampled, and a
human or AI annotator labels their preference according to $P_{R^*}:\Upsilon\times\Upsilon\to\{0,1\}$, where $\Upsilon=\times_H(\mathcal{S}\times\mathcal{A})$ denotes the state-action trajectory space and $P_{R^*}(\tau,\tau')$ indicates whether $\tau$ is preferred to $\tau'$ under the latent oracle reward $R^*: \mathcal{S}^\mathrm{sys} \times \mathcal{A} \to \mathbb{R}$. This yields a preferred trajectory $\tau^w$ and a dispreferred one $\tau^l$. The oracle reward $R^*$ is available to the annotator. Agents learn from the preferences over trajectories. The objective of the Pb-Dec-POMDP is to maximize the expected cumulative latent reward, $\boldsymbol{\pi}^{*}=\arg\max_{\boldsymbol{\pi}}\mathbb{E}_{(s_t, \mathbf{a}_t)\sim\boldsymbol{\pi}}\!\left[\sum_{t=0}^{H-1}R^*(s_t^{\mathrm{sys}},\mathbf{a}_t)\right]$. 

\subsection{Preference-based Centralized MAS}

In addition to decentralized collaboration, a MAS can also coordinate the agents through centralized communication or control. Agents can exchange information from their local histories to construct a shared context, which each agent then acts on~\citep{wu2023autogen,chen2024reconcile}. A manager agent can also aggregate available context about a task and decompose it into jobs for multiple workers~\citep{wu2023autogen,hong2024metagpt}. Such coordination helps agents avoid redundant work or missed subtasks, improving the quality of the joint solution.

This setting can be formulated as a \textit{preference-based multi-agent POMDP} (Pb-MPOMDP), which follows the same tuple as the Pb-Dec-POMDP but adopts a centralized information structure~\citep{messias2011efficient}. Specifically, in Pb-MPOMDP, the policies are conditioned on the full joint action-observation history $\mathbf{h}_t=\{\mathbf{o}_0,\mathbf{a}_0,\ldots,\mathbf{a}_{t-1},\mathbf{o}_t\}$. Each communicative agent $i$ generates its response as $a_{i,t}\sim\pi_i(\cdot|\mathbf{h}_t)$ or a coordinator generates the joint response as $\mathbf{a}_t\sim\boldsymbol{\pi}_{\mathrm{cen}}(\cdot|\mathbf{h}_t)$. 
The environment transition, learning objective, and preference labeling remain unchanged. 

\subsection{Challenges}

\paragraph{Data coverage and density}As it is often difficult and costly to simulate environments involving humans, many preference learning methods adopt pre-collected preference datasets for policy optimization \citep{ziegler2019fine,rafailov2023direct}. However, such datasets may not adequately cover the trajectories induced by the policies, and thus provide insufficient comparisons for effective training \citep{zhan2024provable}. This limitation is amplified in MAS, which induces a combinatorially large policy space \citep{oliehoek2016concise,yang2021believe,shao2023counterfactual}. The lack of data coverage and density can limit effective optimization, as preferences collected from sparse or poorly covered regions may fail to guide learning toward high-quality cooperation policies \citep{zhan2024provable,pan2022plan,barde2024model}. MAPL supports online preference collection, where diverse trajectory comparisons are obtained throughout training.

\paragraph{Multi-agent preference datasets}Although there exist many preference datasets for single LLM alignment \citep{nakano2021webgpt,bai2022training, cui2023ultrafeedback, kopf2023openassistant}, standard datasets tailored to multi-agent collaboration are very limited.
An intuitive approach is to manually adapt existing datasets by assigning roles or subtasks to individual agents. However, many of these tasks cannot be divided into subtasks meaningfully \citep{liu2026llm}. Even when such a division is possible, it remains unclear whether agents can achieve the optimal (or a good) solution under such task decomposition \citep{dietterich2000hierarchical}. MAPL uses joint trajectories of the agents being optimized in the targeted environment to build the dataset. Therefore, the trajectories in the dataset largely preserve the achievable cooperation of these agents.

\paragraph{Accuracy of reward estimates}In cooperative MAS, agents collaborate toward a shared team objective \citep{tan1993multi,claus1998dynamics, oliehoek2016concise,samvelyan2019starcraft, amato2024introduction, albrecht2024multi}. Therefore, it is natural to assign preference to trajectories based on the joint policy. However, unlike MARL with fixed and well-specified rewards, preference-based RL must learn rewards from preferences \citep{christiano2017deep, ziegler2019fine}. With limited or noisy data, such reward estimates can be inaccurate or high-variance, making subsequent value estimation and policy updates less reliable. In addition, when the implicit reward model is updated alongside the policies (e.g., \citealp{rafailov2023direct, guo2024direct}), the training is non-stationary. MAPL maintains a dynamic replay buffer containing diverse samples from different stages of training. We also incorporate a group-relative Monte Carlo-based MARLHF approach (Equation~\ref{eq:marlhf}) and use mini-batch stochastic gradient descent for policy optimization to improve training stability.

\section{Multi-Agent Preference Learning}
\label{sec:mapl_methods}

\begin{figure}[t]
    \centering
    \includegraphics[width=0.88\textwidth]{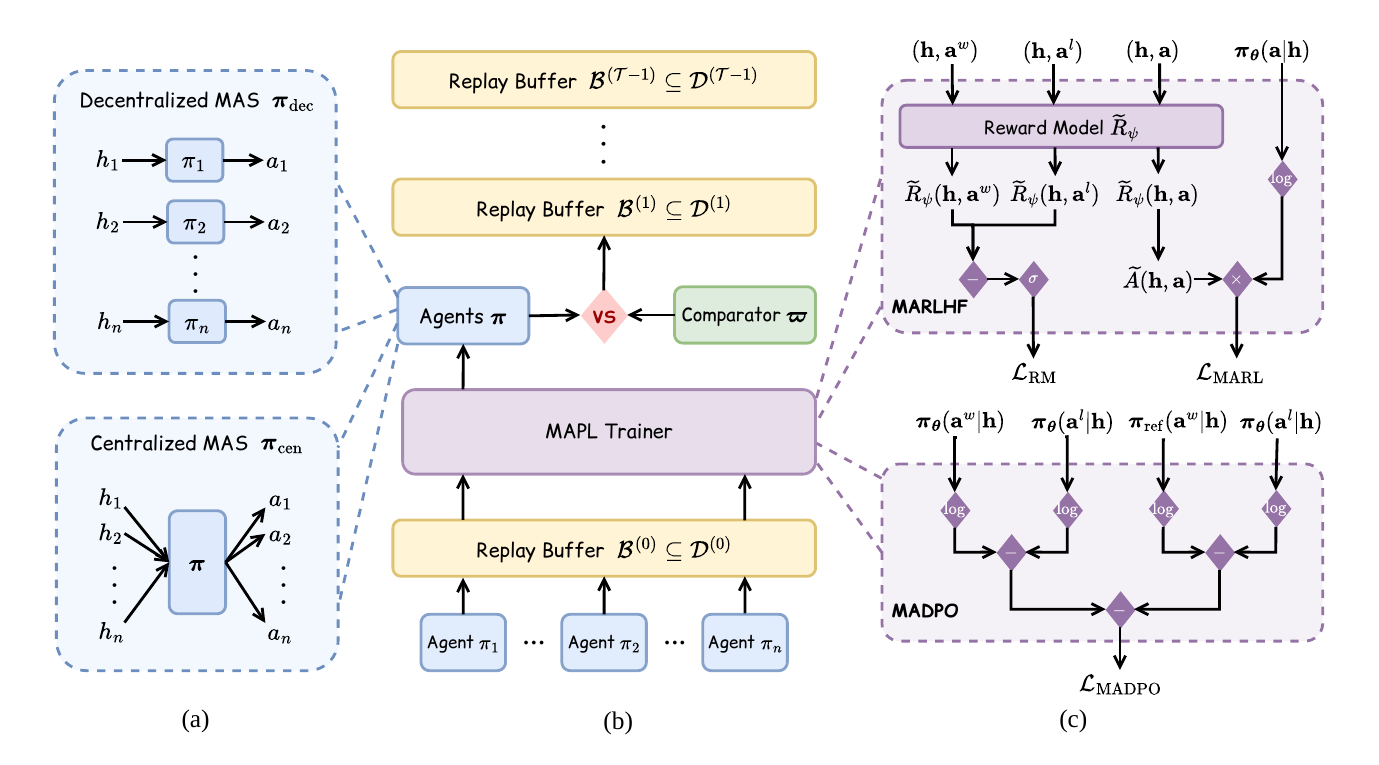}
    \captionsetup{font=footnotesize}
    \vspace{-3mm}
    \caption{An illustration of the MAPL framework and methods. (a) Decentralized and centralized MAS; (b) MAPL framework; (c) Procedures of MARLHF and MADPO.}
    \label{fig:framework}
\end{figure}

We introduce MAPL (Figure~\ref{fig:framework}), a general framework for optimizing cooperative MAS from preference feedback. We instantiate MAPL with two representative approaches, MARLHF and MADPO, and discuss their key design features. Further details and additional MAPL variants are provided in Appendices~\ref{app:derivation} and \ref{app:variants}, respectively.

\subsection{Framework} \label{sec:training_paradigm}

Figure~\ref{fig:framework}(b) illustrates the overall framework of MAPL. We initialize training by collecting the full joint histories $\mathbf{h}_H=\{\mathbf{o}_0,\mathbf{a}_0,\ldots,\mathbf{o}_H\}$ and the corresponding trajectories $\tau_H=\{s_0,\mathbf{a}_0,\ldots,s_H\}$ generated by the LLM agents for each task. For each pair of trajectories $\tau_H$ and $\tau'_H$ of a task, the annotator compares their returns following $R^*$, labels the trajectory with the higher return as the winner $\tau^w$ and the other as the loser $\tau^l$, and discards ties. This yields an initial preference dataset $\mathcal{D}^{(0)}=\{(\mathbf{h}_H^w,\tau_H^w,\mathbf{h}_H^l,\tau_H^l)_j\}_{j=1}^m$ containing $m$ preference pairs, with $\mathbf{h}_H^w$ and $\mathbf{h}_H^l$ denoting the joint histories associated with the winning and losing trajectories, respectively. We then select the top-$k$ preference pairs according to the returns of their winning trajectories to construct the replay buffer $\mathcal{B}^{(0)}\subseteq\mathcal{D}^{(0)}$, which is used for the first iteration of MAPL training.

Since raw LLMs are not inherently cooperative, $\mathcal{D}^{(0)}$ may contain many trajectories with weak or no cooperation. As training progresses, the agents can generate higher-quality cooperative trajectories, providing more informative preference data for subsequent optimization. At each iteration $\nu$, we collect $m$ trajectories $\{(\mathbf{h}_H,\tau_H)_j\}_{j=1}^m\sim\boldsymbol{\pi}$ from the current agent policies and pair them with $m$ trajectories $\{(\mathbf{h}'_H,\tau'_H)_j\}_{j=1}^m\sim\boldsymbol{\varpi}$ generated by a comparator policy, such as previous checkpoints or other model variants. The annotator compares each pair to construct a new preference dataset $\mathcal{D}^{(\nu)}$, where the top-$k$ preference pairs are selected to form the replay buffer $\mathcal{B}^{(\nu)}\subseteq\mathcal{D}^{(\nu)}$. This procedure is repeated until the maximum number of training iterations $\mathcal{T}$ is reached.

The agent policies and the comparator policy can adopt either decentralized or centralized collaboration, as illustrated in Figure~\ref{fig:framework}(a). Under decentralized collaboration, each agent conditions only on its own local history without explicit communication, using $\pi_i(\cdot|h_i)$ or $\varpi_i(\cdot|h'_i)$; under centralized collaboration, the policy conditions on the joint history, using $\boldsymbol{\pi}_{\mathrm{cen}}(\cdot|\mathbf{h})$ or $\boldsymbol{\varpi}_{\mathrm{cen}}(\cdot|\mathbf{h}')$. Since centralized policies have access to global information, they can potentially generate better trajectories with tighter coordination for training. Combining these choices, we therefore consider three training paradigms: \textit{decentralized collaboration with a decentralized comparator} (Dec-Dec), \textit{centralized collaboration with a centralized comparator} (Cen-Cen), and \textit{decentralized collaboration with a centralized comparator} (Dec-Cen).

\subsection{Methods} \label{sec:marlhf}

After obtaining $\mathcal{B}^{(\nu)}$ at iteration $\nu$, we can either learn a reward model and optimize the policies with MARL, or directly optimize the policies from trajectory preferences.

\paragraph{MARLHF}In MARLHF, at each iteration $\nu$, we first learn a history-based reward model $\widetilde{R}_{\psi}(\mathbf{h}_t, \mathbf{a}_t)$ to approximate the expected reward $\mathbb{E}[R^*(s_t^{\mathrm{sys}},\mathbf{a}_t)|\mathbf{h}_t,\mathbf{a}_t]$ for the given joint history. Specifically,
\begin{equation}
\mathcal{L}_{\mathrm{RM}}(\psi)
=
-\mathbb{E}_{(\mathbf{h}_H^{w},\mathbf{h}_H^{l})\sim\mathcal{B}^{(\nu)}}
\left[
\log\sigma\left(
\sum_{t=0}^{H-1}
\left[
\widetilde{R}_{\psi}(\mathbf{h}_t^{w},\mathbf{a}_t^{w})
-
\widetilde{R}_{\psi}(\mathbf{h}_t^{l},\mathbf{a}_t^{l})
\right]
\right)
\right].
\label{eq:marlhf-rm}
\end{equation}
The reward model is then held fixed during policy optimization. Although many MARL algorithms are available for the subsequent policy optimization~\citep{albrecht2024multi,amato2024introduction}, not all are well-suited to this setting. For example, due to the extremely large action space of LLM-based MAS, value-based methods are typically ineffective for LLM policy optimization. Moreover, in practice, the critic may be fit to an inaccurate reward model, especially in the early stage of training. These errors then propagate into the value estimates and undermine policy optimization.

As learning a critic is both difficult and costly, we use a group-based Monte Carlo method for MARL optimization~\citep{liu2026llm}. For each task and policy update, we independently sample $G>1$ trajectories $\tau_H^{\mathcal{G}}=\{\tau_H^{(1)},\ldots,\tau_H^{(G)}\}$ and their corresponding joint histories $\mathbf{h}_H^{\mathcal{G}}=\{\mathbf{h}_H^{(1)},\ldots,\mathbf{h}_H^{(G)}\}$ by executing the current joint policy $\boldsymbol{\pi}_{\mathrm{old}}$. For compactness, we use $\boldsymbol{\pi}_{\boldsymbol{\theta}}(\mathbf{h}_H)=\prod_{t=0}^{H-1}\boldsymbol{\pi}_{\boldsymbol{\theta}}(\mathbf{a}_t|\mathbf{h}_t)$ to denote the product of joint action probabilities along the history $\mathbf{h}_H$. We adopt the averaged prediction within each group as a baseline to construct the group-relative advantage $\widetilde{A}^{(g)}$ and optimize the agents according to,
\begin{equation}
\mathcal{L}_{\mathrm{MARL}^{\mathcal{G}}}(\boldsymbol{\theta})
=
-\mathbb{E}_{\mathbf{h}_{H}^{\mathcal{G}}\sim\boldsymbol{\pi}_{\mathrm{old}}}
\left[
\frac{1}{G}\sum_{g=1}^{G}
\mathsf{sg}\!\left[
\frac{G}{G-1}\widetilde{A}^{(g)}
-\eta\log\frac{
\boldsymbol{\pi}_{\boldsymbol{\theta}}(\mathbf{h}_{H}^{(g)})
}{
\boldsymbol{\pi}_{\mathrm{ref}}(\mathbf{h}_{H}^{(g)})
}
\right]
\log\boldsymbol{\pi}_{\boldsymbol{\theta}}(\mathbf{h}_{H}^{(g)})
\right].
\label{eq:marlhf}
\end{equation}
where $\widetilde{A}^{(g)}=\sum_{t=0}^{H-1}
\left[
\widetilde{R}_{\psi}(\mathbf{h}_t^{(g)},\mathbf{a}_t^{(g)})
-\frac{1}{G}\sum_{g'=1}^{G}
\widetilde{R}_{\psi}(\mathbf{h}_t^{(g')},\mathbf{a}_t^{(g')})
\right]$,
$\mathsf{sg}$ denotes the stop-gradient operator. $\frac{G}{G-1}$ corrects the group average, and the KL term includes the log-ratio as a fixed weight to produce the trajectory-level KL gradient in expectation.
The agents' policies $\pi_{\theta_i}$ and $\boldsymbol{\pi}_{\boldsymbol{\theta}}$ are then updated as $\nabla_{\theta_i}\mathcal{L}_{\mathrm{MARL}^{\mathcal{G}}}(\boldsymbol{\theta})$ and $\nabla_{\boldsymbol{\theta}}\mathcal{L}_{\mathrm{MARL}^{\mathcal{G}}}(\boldsymbol{\theta})$ in decentralized and centralized MAS, respectively.

\paragraph{MADPO}We also present a more efficient approach that directly optimizes the policy from
$\mathcal{B}^{(\nu)}$. At each iteration $\nu$, MADPO minimizes
\begin{equation}
\mathcal{L}_{\mathrm{MADPO}}(\boldsymbol{\theta})
=
-\mathbb{E}_{(\mathbf{h}_H^w,\mathbf{h}_H^l)\sim\mathcal{B}^{(\nu)}}
\left[
\log\sigma\left(
\eta\log\frac{\boldsymbol{\pi}_{\boldsymbol{\theta}}(\mathbf{h}_H^w)}{\boldsymbol{\pi}_{\mathrm{ref}}(\mathbf{h}_H^w)}
-
\eta\log\frac{\boldsymbol{\pi}_{\boldsymbol{\theta}}(\mathbf{h}_H^l)}{\boldsymbol{\pi}_{\mathrm{ref}}(\mathbf{h}_H^l)}
\right)
\right].
\label{eq:madpo}
\end{equation}
MADPO implicitly parameterizes the reward as
$\widetilde{R}_{\boldsymbol{\theta}}(\mathbf{h}_t,\mathbf{a}_t)=\eta\log\dfrac{\boldsymbol{\pi}_{\boldsymbol{\theta}}(\mathbf{a}_t|\mathbf{h}_t)}{\boldsymbol{\pi}_{\mathrm{ref}}(\mathbf{a}_t|\mathbf{h}_t)}$. Then we can minimize the loss in Equation~\ref{eq:madpo} with respect to each agent's parameters $\theta_i$ under decentralized collaboration, or the coordinator's parameters $\boldsymbol{\theta}$ under centralized coordination. 

We provide further details about MAPL derivations and algorithms in Appendix~\ref{app:derivation}.

\section{Experiments} \label{sec:experiments}

We evaluate MAPL in four domains: \texttt{TLDR} for document processing \citep{volske2017tl}, \texttt{CoopHE} for programming \citep{liu2026llm}, \texttt{BFCL} for function calling \citep{patil2025berkeley}, and \texttt{Travel} for travel planning \citep{xie2024travelplanner}. Experimental settings, additional results, prompt design, and compute resources are provided in Appendices~\ref{app:exp_setting}, \ref{app:additional_results}, \ref{app:prompt}, and \ref{app:resources}, respectively.

\subsection{Setup}

As discussed in \S~\ref{sec:background}, existing large-scale preference datasets~\citep{bai2022training,cui2023ultrafeedback} cannot be seamlessly repurposed for MAPL, since many tasks are atomic and cannot induce meaningful collaboration. Following prior work~\citep{christiano2017deep,lee2021pebble,kim2023preference,zhu2024decoding}, we use deterministic oracle reward functions to generate preferences (Appendix~\ref{app:dataset_details}).

\paragraph{TLDR}Extracting key information from long context is an important ability for LLM agents. We instantiate this setting as a text summarization task built on the \textit{TL;DR} dataset~\citep{volske2017tl}. In \texttt{TLDR}, two \textit{Qwen3-1.7B} agents collaborate to summarize Reddit posts provided in the \textit{prompt} field. One agent produces a concise high-level summary, while the other provides a detailed summary with more elaborated information. Their outputs are merged into a single briefing presented to the reader. We measure summarization quality with a weighted score over structural quality, stylistic consistency, and logical coherence, with details provided in Appendix~\ref{app:dataset_details}.

\paragraph{CoopHE}Prior work has shown the promise of employing multiple LLMs for software development~\citep{talebirad2023multi,qian2024chatdev}. We take the collaborative code generation tasks in the \textit{CoopHumanEval} \citep{liu2026llm} dataset as a representative case. In \texttt{CoopHE}, two \textit{Qwen2.5-Coder-3B} coding agents are solving basic \textit{Python} programming problems, each specified by a natural-language prompt in \textit{prompt} and a target function signature in \textit{entry\_point}. One agent implements utility modules in \textit{aux}($\cdot$), while the other constructs \textit{main}($\cdot$) by invoking them. The two functions are aggregated into a single file as a solution. We report the test pass rate for functional correctness and the valid auxiliary call rate for collaboration.

\paragraph{BFCL}Modern LLM agents interact with the systems through diverse tools (e.g., MCP servers, shell toolkits, and browser interfaces)~\citep{anthropic2024mcp, yang2024swe,zhou2024webarena}, yet sequentially invoking these tools incurs substantial latency. Multi-agent collaboration mitigates this by issuing function calls in parallel. We use the \textit{parallel} and \textit{parallel\_multiple} tasks in the \textit{Berkeley Function Calling Leaderboard} (\texttt{BFCL}) to demonstrate this concept~\citep{patil2025berkeley}. Two \textit{Qwen3-4B-Instruct-2507} agents are jointly generating structured function calls for the request in \textit{user\_prompt}, following the schemas in \textit{function}. In \textit{parallel} tasks, the agents share a single function schema but may call it with different arguments, whereas in \textit{parallel\_multiple} tasks, they must select the appropriate function from several available schemas. Agents are expected to learn a cooperation scheme that balances workloads and produces complementary calls. We use functional correctness and costs (\#calls) as evaluation metrics.

\paragraph{Travel}We also consider a more concrete agentic domain, where two \textit{Qwen3-4B-Instruct-2507} agents work together to plan a short trip. We draw tasks from \textit{TravelPlanner}~\citep{xie2024travelplanner}. We select the ones with $\textit{days}\leqslant$ 5 and $\textit{visiting\_city\_number}\leqslant$ 2 for simplicity and finer-grained comparison. The full request is provided in \textit{query}, while candidate itinerary options and their associated costs are given in \textit{reference\_information}. In \texttt{Travel}, one agent arranges logistics (e.g., the city route, transportation, and accommodation), while the other designs activities (e.g., restaurants and attractions). Their outputs are combined into a single itinerary. We report a coarse-grained and a fine-grained metric: \textit{pass} measures the proportion of plans satisfying all commonsense constraints (e.g., consistency between scheduled activities and the city occupied on that day), whereas \textit{success} additionally requires compliance with the budget limits in \textit{budget} and the user-specified constraints in \textit{local\_constraint}, including room types, food and transportation preferences.

\begin{figure}[t]
    \centering
    \includegraphics[width=0.88\textwidth]{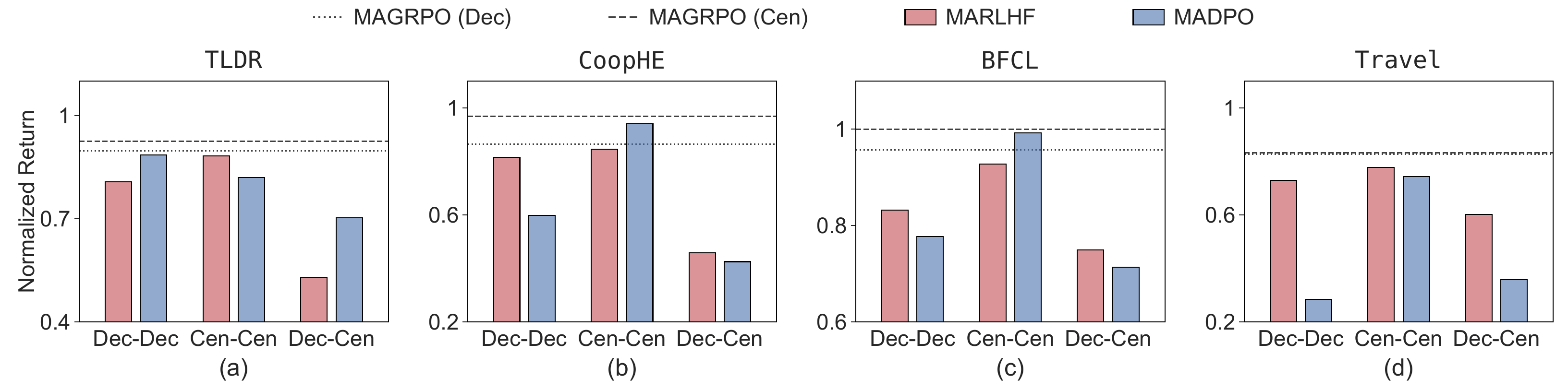}
    \captionsetup{font=footnotesize}
    \vspace{-3mm}
    \caption{A comparison of MARLHF (\textcolor{red}{red}) and MADPO (\textcolor{blue}{blue}) under Dec-Dec, Cen-Cen and Dec-Cen on \texttt{TLDR}, \texttt{CoopHE}, \texttt{BFCL}, and \texttt{Travel} across 5 runs. Dotted and dashed lines denote the performance of MAGRPO on decentralized and centralized MAS. Results are normalized to the reward scale w.r.t. each domain.}
    \label{fig:main-results}
\end{figure}

\paragraph{Baselines}We compare our methods against three categories of baselines: single-agent methods, test-time LLM collaboration frameworks, and MARL approaches, with detailed prompts in Appendix~\ref{app:prompt}. Single-agent baselines employ a single LLM comparable in size to the MAS. We include the base model and its fine-tuned variants by RL and preference learning algorithms. For test-time collaboration, we consider three representative frameworks: \textit{parallel}, where agents execute concurrently without communication; \textit{sequential}, where one agent executes after another and communicates one-way~\citep{wu2023autogen}; and \textit{discussion}, where agents engage in bidirectional communication~\citep{du2023improving}. Given the prohibitively large policy space of LLMs, value-based methods are often impractical. We thus consider policy gradient methods as MARL baselines: multi-agent GRPO (MAGRPO)~\citep{liu2026llm}, which extends GRPO to the multi-agent setting with group-relative advantages, and collaborative-LLM actor-critic with a centralized critic (CoLLM-CC) and with decentralized critics (CoLLM-DC)~\citep{liu2026learning}.

\subsection{Experimental Results}

\begin{wraptable}{r}{0.6\textwidth}
\centering
\setlength{\tabcolsep}{2pt}
\renewcommand{\arraystretch}{1.15}
\resizebox{0.98\linewidth}{!}{
\begin{tabular}{l@{\hspace{3pt}}ccccccc}
\toprule
\multirow{2}{*}[-0.5ex]{\textbf{Method}}
& \multicolumn{1}{c}{\textbf{TLDR}}
& \multicolumn{2}{c}{\textbf{CoopHE}}
& \multicolumn{2}{c}{\textbf{BFCL}}
& \multicolumn{2}{c}{\textbf{Travel}} \\
\cmidrule(lr){2-2}
\cmidrule(lr){3-4}
\cmidrule(lr){5-6}
\cmidrule(lr){7-8}
& Score
& Pass
& Call
& Correct
& Costs
& Pass
& Success \\
\midrule

\rowcolor{pptint}
\multicolumn{8}{l}{\textbf{\textit{(a) Single Agent}}} \\
Raw Model
& 30.3 & 56.3 & \textit{N/A} & 47.5 & 2.4 & 71.3 & 9.6 \\
GRPO~\citep{shao2024deepseekmath}
& 91.7 & 61.8 & \textit{N/A} & 50.2 & 2.5 & 74.5 & \textbf{25.0} \\
PPO~\citep{schulman2017proximal}
& 94.5 & 62.5 & \textit{N/A} & \textbf{56.3} & 2.5 & 73.2 & 14.5 \\
RLHF~\citep{christiano2017deep}
& 79.4 & 65.0 & \textit{N/A} & 48.3 & 2.3 & 75.6 & 5.6 \\
DPO~\citep{rafailov2023direct}
& 88.0 & 44.2 & \textit{N/A} & 49.2 & 2.4 & 59.8 & 3.8 \\

\midrule
\rowcolor{pptint}
\multicolumn{8}{l}{\textbf{\textit{(b) Test-Time Collaboration}}} \\
Parallel~\citep{wu2023autogen}
& 22.9 & 50.0 & 18.4 & 12.8 & 1.2 & 66.2 & 8.3 \\
Sequential~\citep{wu2023autogen}
& 21.7 & 62.5 & 24.6 & 25.5 & 1.3 & 66.9 & 10.0 \\
Discussion~\citep{du2023improving}
& 22.3 & 25.0 & 12.4 & 9.8 & 1.4 & 62.5 & 4.2 \\

\midrule
\rowcolor{pptint}
\multicolumn{8}{l}{\textbf{\textit{(c) MARL with $R^\ast$}}} \\
MAGRPO~\citep{liu2026llm}
& 93.5 & 71.3 & 37.9 & 43.8 & 1.0 & 76.2 & 12.5 \\
CoLLM-DC~\citep{liu2026learning}
& \textbf{95.4} & 59.3 & 23.2 & 28.7 & \textbf{0.9} & 42.8 & 0.0 \\
CoLLM-CC~\citep{liu2026learning}
& 95.2 & \textbf{75.2} & 39.7 & 53.4 & 1.1 & \textbf{84.3} & \textbf{25.0} \\

\midrule
\rowcolor{pptint}
\multicolumn{8}{l}{\textbf{\textit{(d) MAPL (Ours)}}} \\
MARLHF (Dec-Dec)$^\dagger$
& 86.3
& 61.0
& 40.2
& 22.7
& \underline{1.0}
& 70.5
& 13.1 \\
MARLHF (Cen-Cen)
& 92.0
& 64.4
& \textbf{\underline{46.8}}
& 38.5
& 2.5
& \underline{76.4}
& \underline{14.5} \\
MARLHF (Dec-Cen)
& 56.4
& 20.3
& 14.3
& 14.9
& 1.3
& 62.4
& 8.3 \\

\hdashline

MADPO (Dec-Dec)$^\ddagger$
& \underline{92.9}
& 58.8
& 28.5
& 15.6
& 1.1
& 41.5
& 0.0 \\
MADPO (Cen-Cen)
& 87.4
& \underline{65.8}
& 44.9
& \underline{48.3}
& 2.5
& 73.6
& 12.5 \\
MADPO (Dec-Cen)
& 74.9
& 16.5
& 12.2
& 11.1
& \underline{1.0}
& 40.6
& 4.2 \\

\bottomrule
\end{tabular}
}
\captionsetup{font=footnotesize}
\vspace{-2mm}
\caption{
Evaluation results of MAPL and baseline methods on \texttt{TLDR}, \texttt{CoopHE}, \texttt{BFCL}, and \texttt{Travel} across 5 runs: Score measures the summary quality; Pass denotes the pass rate; Call denotes the invocation rate of \textit{aux($\cdot$)}; Correct is the functional correctness rate; Costs measures the number of calls per agent; Pass is the commonsense pass rate; Success is the final pass rate with more constraints. \underline{Underline} and \textbf{bold} indicate the best results across MAPL and all methods. $^\dagger$ and $^\ddagger$ denote pivots for the ablation studies in Table~\ref{tab:main-ablation}.
}
\label{tab:main_results}
\end{wraptable}
We compare MAPL with single-agent, test-time collaboration, and oracle-reward MARL baselines. We set $G=4$ in Equation~\ref{eq:marlhf} and draw comparators from the same model family as the agents. The decentralized comparator uses the same model as each agent, whereas the centralized comparator approximately matches the agents' combined parameter count for a fair comparison (Appendix~\ref{app:exp_setting}).

As shown in Table~\ref{tab:main_results} and Figure~\ref{fig:main-results}, MAPL under the best configurations can reach pass rates of 65.8\% on \texttt{CoopHE} and 76.4\% on \texttt{Travel}, exceeding the strongest single-agent baseline on \texttt{CoopHE} and \texttt{Travel} and outperforming test-time collaboration across all four domains. MARL methods attain the highest \texttt{TLDR} score and pass rates on \texttt{CoopHE} and \texttt{Travel}, since oracle rewards align policy optimization directly with the task objectives. Nevertheless, MAPL closely approaches MAGRPO, with less than 1\% difference in \texttt{TLDR} score and \texttt{Travel} pass rate. Surprisingly, we find MARLHF can even surpass MAGRPO on \texttt{CoopHE} call rate (40.2\% vs. 37.9\%) and \texttt{Travel} success (13.1\% vs. 12.5\%) under decentralized collaboration. We attribute this to a higher density of samples of similar quality in the MAPL preference dataset (e.g., correct code with different cooperation patterns), which could provide richer local training signals for distinguishing their relative quality.

Among the MAPL paradigms, Cen-Cen generally performs best for both MARLHF and MADPO across most domains, as centralized agents can exploit more complete information to coordinate. Dec-Dec also achieves strong performance after proper tuning and can outperform the raw larger model on \texttt{CoopHE}. This demonstrates the potential of decentralized collaboration, where multiple smaller models can cooperate to surpass a larger one. Cen-Cen underperforms Dec-Dec with MADPO on \texttt{TLDR}, where both the centralized agent and the comparator use a relatively small model compared with those in other domains. This suggests that centralized collaboration may require greater model capacity than individual decentralized agents, as centralized information typically induces a longer context for model reasoning. Dec-Cen generally performs worse due to the information and realizability gap between the comparator and agents. We find that this decrease is more salient in MARLHF than in MADPO, as its two-stage pipeline amplifies this representation error.

\subsection{Ablations} \label{sec:experiments_ablation}

We ablate MAPL on replay buffers, comparators, and MARLHF methods. We use the Dec-Dec paradigm as the representative setting and provide additional results in Appendix~\ref{app:additional_results}.

\begin{wraptable}{r}{0.6\textwidth}
\centering
\setlength{\tabcolsep}{8pt}
\renewcommand{\arraystretch}{1.2}
\resizebox{0.98\linewidth}{!}{
\begin{tabular}{l@{\hspace{10pt}}cccc}
\toprule
\multirow{2}{*}[-0.5ex]{\textbf{Method}}
& \multicolumn{1}{c}{\textbf{TLDR}}
& \multicolumn{1}{c}{\textbf{CoopHE}}
& \multicolumn{1}{c}{\textbf{BFCL}}
& \multicolumn{1}{c}{\textbf{Travel}} \\
\cmidrule(lr){2-2}
\cmidrule(lr){3-3}
\cmidrule(lr){4-4}
\cmidrule(lr){5-5}
& Score
& Pass
& Correct
& Pass \\
\midrule

\rowcolor{pptint}
\multicolumn{5}{l}{\textit{\textbf{(a) Replay Buffer}}} \\
MARLHF\, ($\lambda$-Decay)$^\dagger$
& \underline{86.3}
& \textbf{\underline{61.0}}
& \textbf{\underline{22.7}}
& \textbf{\underline{70.5}} \\
MARLHF\, (Offline)
& 70.9 \textcolor{blue}{\scriptsize($\downarrow$15.4)}
& 46.4 \textcolor{blue}{\scriptsize($\downarrow$14.6)}
& 7.5 \textcolor{blue}{\scriptsize($\downarrow$15.2)}
& 63.6 \textcolor{blue}{\scriptsize($\downarrow$6.9)} \\
MARLHF\, (Online)
& 76.0 \textcolor{blue}{\scriptsize($\downarrow$10.3)}
& 59.7 \textcolor{blue}{\scriptsize($\downarrow$1.3)}
& 14.8 \textcolor{blue}{\scriptsize($\downarrow$7.9)}
& 69.8 \textcolor{blue}{\scriptsize($\downarrow$0.7)} \\

\hdashline

MADPO\, ($\lambda$-Decay)$^\ddagger$
& \textbf{\underline{92.9}}
& \underline{58.8}
& \underline{15.6}
& 41.5 \\
MADPO\, (Offline)
& 87.1 \textcolor{blue}{\scriptsize($\downarrow$5.8)}
& 23.0 \textcolor{blue}{\scriptsize($\downarrow$35.8)}
& 3.1 \textcolor{blue}{\scriptsize($\downarrow$12.5)}
& \underline{46.8} \textcolor{red}{\scriptsize($\uparrow$5.3)} \\
MADPO\, (Online)
& 88.7 \textcolor{blue}{\scriptsize($\downarrow$4.2)}
& 58.5 \textcolor{blue}{\scriptsize($\downarrow$0.3)}
& 15.3 \textcolor{blue}{\scriptsize($\downarrow$0.3)}
& 43.2 \textcolor{red}{\scriptsize($\uparrow$1.7)} \\

\midrule

\rowcolor{pptint}
\multicolumn{5}{l}{\textit{\textbf{(b) Comparator}}} \\
MARLHF\, (Same-size)$^\dagger$
& \underline{86.3}
& 61.0
& 22.7
& 70.5 \\
MARLHF\, ($2\times$)
& 66.9 \textcolor{blue}{\scriptsize($\downarrow$19.4)}
& 48.8 \textcolor{blue}{\scriptsize($\downarrow$12.2)}
& 18.8 \textcolor{blue}{\scriptsize($\downarrow$3.9)}
& 56.4 \textcolor{blue}{\scriptsize($\downarrow$14.1)} \\
MARLHF\, ($4\times$)
& 66.5 \textcolor{blue}{\scriptsize($\downarrow$19.8)}
& 50.0 \textcolor{blue}{\scriptsize($\downarrow$11.0)}
& \textbf{\underline{22.9}} \textcolor{red}{\scriptsize($\uparrow$0.2)}
& 65.6 \textcolor{blue}{\scriptsize($\downarrow$4.9)} \\
MARLHF\, (\textit{Deepseek-Pro})
& 70.1 \textcolor{blue}{\scriptsize($\downarrow$16.2)}
& \underline{63.5} \textcolor{red}{\scriptsize($\uparrow$2.5)}
& 21.8 \textcolor{blue}{\scriptsize($\downarrow$0.9)}
& \textbf{\underline{71.2}} \textcolor{red}{\scriptsize($\uparrow$0.7)} \\

\hdashline

MADPO\, (Same-size)$^\ddagger$
& \textbf{\underline{92.9}}
& 58.8
& 15.6
& 41.5 \\
MADPO\, ($2\times$)
& 61.3 \textcolor{blue}{\scriptsize($\downarrow$31.6)}
& 49.2 \textcolor{blue}{\scriptsize($\downarrow$9.6)}
& 3.8 \textcolor{blue}{\scriptsize($\downarrow$11.8)}
& 43.2 \textcolor{red}{\scriptsize($\uparrow$1.7)} \\
MADPO\, ($4\times$)
& 71.8 \textcolor{blue}{\scriptsize($\downarrow$21.1)}
& 49.8 \textcolor{blue}{\scriptsize($\downarrow$9.0)}
& 21.9 \textcolor{red}{\scriptsize($\uparrow$6.3)}
& 46.9 \textcolor{red}{\scriptsize($\uparrow$5.4)} \\
MADPO\, (\textit{Deepseek-Pro})
& 74.6 \textcolor{blue}{\scriptsize($\downarrow$18.3)}
& \textbf{\underline{65.7}} \textcolor{red}{\scriptsize($\uparrow$6.9)}
& \underline{22.7} \textcolor{red}{\scriptsize($\uparrow$7.1)}
& \underline{47.4} \textcolor{red}{\scriptsize($\uparrow$5.9)} \\

\midrule

\rowcolor{pptint}
\multicolumn{5}{l}{\textit{\textbf{(c) MARLHF Method}}} \\
MARLHF\, (MAGRPO)$^\dagger$
& \textbf{\underline{86.3}}
& \textbf{\underline{61.0}}
& \textbf{\underline{22.7}}
& \textbf{\underline{70.5}} \\
MARLHF\, (CoLLM-CC)
& 23.9 \textcolor{blue}{\scriptsize($\downarrow$62.4)}
& 46.5 \textcolor{blue}{\scriptsize($\downarrow$14.5)}
& 7.2 \textcolor{blue}{\scriptsize($\downarrow$15.5)}
& 37.5 \textcolor{blue}{\scriptsize($\downarrow$33.0)} \\
MARLHF\, (CoLLM-DC)
& 26.7 \textcolor{blue}{\scriptsize($\downarrow$59.6)}
& 38.9 \textcolor{blue}{\scriptsize($\downarrow$22.1)}
& 4.6 \textcolor{blue}{\scriptsize($\downarrow$18.1)}
& 14.6 \textcolor{blue}{\scriptsize($\downarrow$55.9)} \\

\bottomrule
\end{tabular}
}
\captionsetup{font=footnotesize}
\vspace{-2mm}
\caption{Ablation results on (a) replay buffers, (b) comparators, and (c) MARLHF methods under Dec-Dec. Pivot entries $^\dagger$ and $^\ddagger$ reported in Table~\ref{tab:main_results} use the $\lambda$-decay buffer, agent-size comparator, and MAGRPO. Parenthesized values show absolute changes relative to the corresponding pivot, where \textcolor{red}{$\uparrow$} denotes an increase and \textcolor{blue}{$\downarrow$} denotes a decrease. Best results within each ablation are reported in \textbf{bold}, best results for each method within each ablation are
\underline{underlined}.
}
\label{tab:main-ablation}
\end{wraptable}
We compare three ways of constructing the replay buffer. Offline MAPL uses 80 pre-collected preference pairs and selects the 16 highest-return samples. Online MAPL collects the same total amount of data over four iterations, where we append the 4 best samples from 20 preference pairs to the buffer at each iteration. We also consider a $\lambda$-decay replay buffer (the default in Table~\ref{tab:main_results}), which downweights samples from earlier iterations with ratio $\lambda$. We use tuned hyperparameter $\lambda\!=\!0.8$ for MARLHF and $\lambda\!=\!0.2$ for MADPO. Table~\ref{tab:main-ablation}(a) shows that online MAPL outperforms offline MAPL on \texttt{TLDR}, \texttt{CoopHE}, and \texttt{BFCL}. For example, MARLHF improves from 7.5\% to 14.8\%, and MADPO from 3.1\% to 15.3\% on \texttt{BFCL}. MAPL with $\lambda$-decay buffer achieves the best results across most domains. We find this improvement to be more pronounced for MARLHF than MADPO, since the broader data coverage supports learning a reward model with more diverse representations.

Then we compare same-family comparators (\textit{Qwen2.5} or \textit{Qwen3}) at $1\times$, $2\times$, and $4\times$ the agent-model size, and \textit{DeepSeek-V4-Pro} from a different model family. Table~\ref{tab:main-ablation}(b) shows that scaling the comparator does not consistently improve performance. On \texttt{TLDR}, scaling from same-size to $4\times$ reduces MARLHF from 86.3\% to 66.5\% and MADPO from 92.9\% to 71.8\%; \textit{DeepSeek-V4-Pro} achieves 70.1\% and 74.6\%, respectively. This is due to the capacity gap between comparators and target agents. Despite receiving the same partial information under Dec-Dec, smaller agents may struggle to realize behaviors favored by stronger comparators. This gap also varies for different optimization methods. With \textit{DeepSeek-V4-Pro}, MADPO improves over the same-size setting from 58.8\% to 65.7\% on \texttt{CoopHE} and from 15.6\% to 22.7\% on \texttt{BFCL}, whereas MARLHF changes from 61.0\% to 63.5\% and from 22.7\% to 21.8\%, respectively. On \texttt{Travel}, MADPO improves from 41.5\% to 47.4\%, compared with 70.5\% to 71.2\% for MARLHF. These results suggest that MADPO benefits more from this external comparator than MARLHF in these domains.

\begin{figure}[t]
    \centering
    \includegraphics[width=0.88\textwidth]{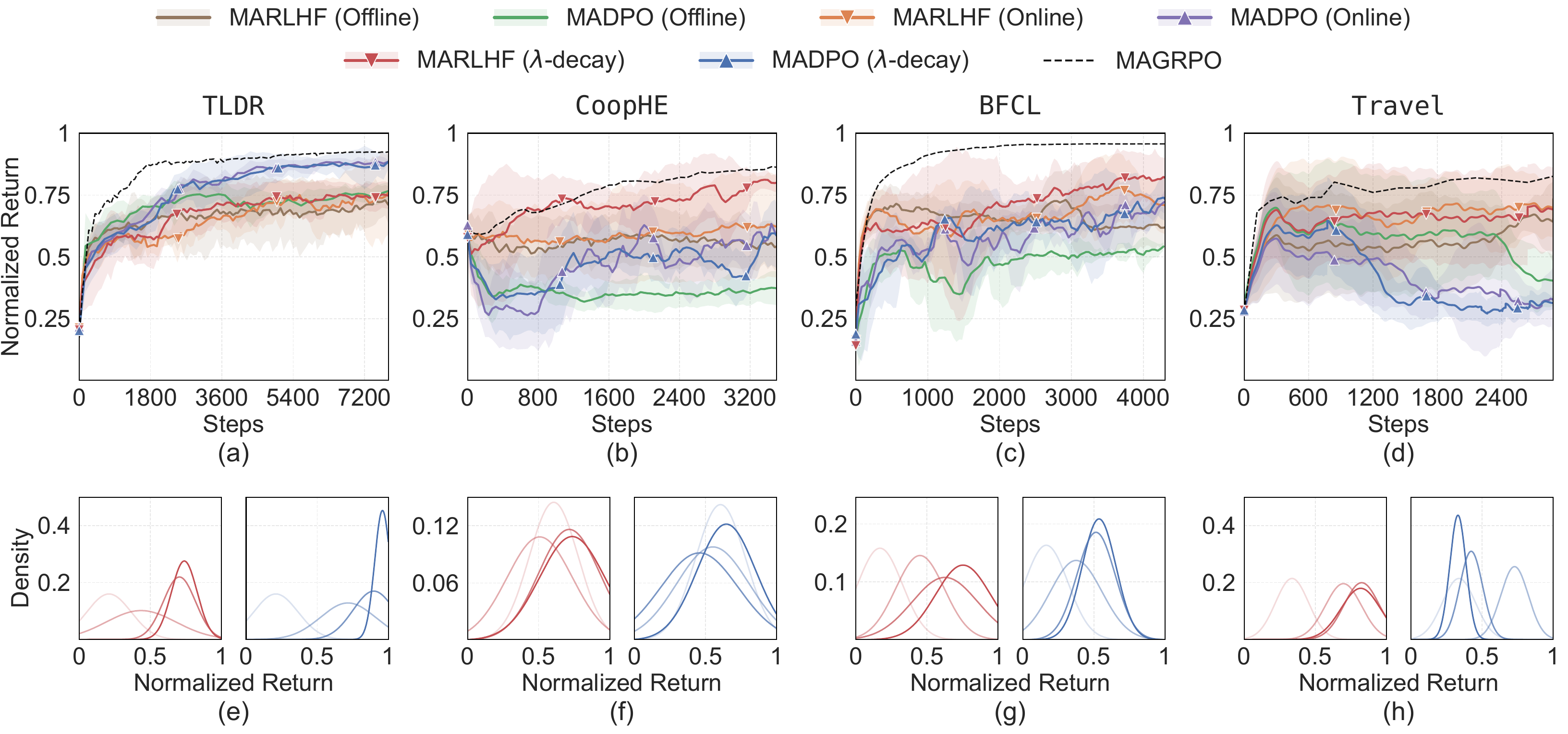}
    \captionsetup{font=footnotesize}
    \vspace{-2mm}
    \caption{Evaluation results (a-d) and data distribution shift (e-h) of MARLHF and MADPO on offline (\textcolor{brown}{brown} and \textcolor{green}{green}), online (\textcolor{orange}{orange} and \textcolor{purple}{purple}), $\lambda$-decay (\textcolor{red}{red} and \textcolor{blue}{blue}) replay buffers. Curves are smoothed by an exponential moving average, and shaded regions indicate the standard deviation. Triangles mark the start of iterations. Data distributions are fitted with a Gaussian. Curves with higher opacity indicate later iterations.}
    \label{fig:main-ablation}
\end{figure}
Finally, we compare MAGRPO with CoLLM-CC and CoLLM-DC, as MARL algorithms for MARLHF. Table~\ref{tab:main-ablation}(c) shows that MAGRPO outperforms both alternatives across all four domains, improving \texttt{TLDR} score by at least 59.6\% and \texttt{CoopHE} pass rate, \texttt{BFCL} correct rate, and \texttt{Travel} pass by at least 14.5\%, 15.5\%, and 33.0\%, respectively. We also observe less stable training with the actor-critic variants, as fitting critics to imperfect or evolving rewards may introduce additional approximation error and further destabilize policy optimization.

\section{Conclusion}

We studied preference-based optimization of cooperative LLM systems. We introduced MAPL, a general framework that iteratively updates agent policies by comparing their solutions with those generated by a comparator policy. We instantiated MAPL with MARLHF and MADPO, and evaluated them under the Dec-Dec, Cen-Cen, and Dec-Cen paradigms. Results on collaborative writing, coding, tool use, and travel planning show that MAPL improves both decentralized collaboration and centralized collaboration, and approaches the performance of MARL with well-specified rewards. Our ablations further show that a larger or centralized comparator does not necessarily facilitate training, due to the information and realizability gaps between agents and comparators.





\section*{Acknowledgments}

This work was partially funded by NSF grants \#2409351 and \#2525087. This work was completed in part using the Explorer Cluster, supported by Northeastern University's Research Computing team. The authors also acknowledge the MGHPCC AI Computing Resource (AICR), with support from the Commonwealth of Massachusetts and the AICR participating institutions, for providing AI and high-performance computing resources used in the results reported in this publication. Shuo Liu received computing support from Lambda's Research Grant Program. We thank members of the Lab for Learning and Planning in Robotics for valuable discussion.

\clearpage
\bibliography{reference}
\bibliographystyle{iclr2027_conference}

\clearpage
\appendix
\pagestyle{appendix}
\startcontents[appendix]
{
\hypersetup{linkcolor=ppappref}
\section*{Appendix Contents}
\printcontents[appendix]{}{1}{
  \setcounter{tocdepth}{2}
}
}
\clearpage
\hypersetup{linkcolor=ppref}
\input{appendix/derivation}
\input{appendix/variants}

\input{appendix/exp-setting}
\input{appendix/additional}
\input{appendix/prompt}
\input{appendix/compute}
\input{appendix/related}
\input{appendix/impact}

\end{document}

%% file: math_commands.tex
\usepackage{amsmath,amsfonts,bm}

\def\eqref#1{equation~\ref{#1}}

\def\1{\bm{1}}

\DeclareMathAlphabet{\mathsfit}{\encodingdefault}{\sfdefault}{m}{sl}
\SetMathAlphabet{\mathsfit}{bold}{\encodingdefault}{\sfdefault}{bx}{n}



%% file: appendix/derivation.tex
\section{MAPL Details} \label{app:derivation}

\subsection{MAPL Algorithms}

We present MARLHF and MADPO under Dec-Dec in Algorithms~\ref{alg:mapl} and~\ref{alg:madpo} as representatives. Both methods follow the same preference collection procedure. We obtain Dec-Cen by replacing the decentralized comparators with a centralized comparator $\boldsymbol{\varpi}_{\mathrm{cen}}(\cdot|\mathbf{h}'_t)$ conditioned on the joint history. For Cen-Cen, we also replace the decentralized agents with a centralized coordinator. The response-generation block in Algorithm~\ref{alg:mapl}, also used by Algorithm~\ref{alg:madpo}, then generates joint responses $\mathbf{a}_t\sim \boldsymbol{\pi}_{\boldsymbol{\theta}}(\cdot|\mathbf{h}_t)$. The policy-update blocks in both algorithms apply an update with respect to $\boldsymbol{\theta}$.

\begin{algorithm}[h]
\caption{MARLHF under Dec-Dec}
\label{alg:mapl}
\small
\begin{algorithmic}[1]
\State \textbf{Input:} Task dataset $\mathcal{X}$;
LLM agents $\{\pi_{\theta_i}\}_{i\in\mathcal{I}}$;
LLM comparators $\{\varpi_i\}_{i\in\mathcal{I}}$;
LLM reference policy $\boldsymbol{\pi}_{\mathrm{ref}}$;
human annotator following oracle reward $R^*$;
learning rates $\alpha_\pi,\alpha_R$;
training batch size $b$;
pair count $m$; buffer size $k$; group size $G$;
horizon $H$; iterations $\mathcal{T}$; KL coefficient $\eta>0$.
\For{iteration $\nu = 0,\ldots,\mathcal{T}-1$}
    \State $\mathcal{D}^{(\nu)} \gets \emptyset$.
    \For{each task $x\sim\mathcal{X}$
    until $|\mathcal{D}^{(\nu)}|=m$}
        \State Initialize task prompts $o_{i,0}$ and
        histories $h_{i,0}\gets \{o_{i,0}\}$,
        $\forall i\in\mathcal{I}$.
        \label{line:mapl-rollout-start}
        \For{turn $t=0,\ldots,H-1$}
            \BeginBox[draw=gray, line width=0.6pt]
                \For{each agent $i\in\mathcal{I}$ in parallel}
                    \State Generate response
                    $a_{i,t}\sim
                    \pi_{\theta_i}(\cdot| h_{i,t})$.
                \EndFor
            \EndBox
            \State Aggregate responses
            $\mathbf{a}_t=\{a_{i,t}\}_{i\in\mathcal{I}}$.
            \State Receive new prompts $o_{i,t+1}$ and update
            $h_{i,t+1}\gets\{h_{i,t},a_{i,t},o_{i,t+1}\}$,
            $\forall i\in\mathcal{I}$.
        \EndFor
        \State Record joint history $\mathbf{h}_H$
        and trajectory $\tau_H$.
        \label{line:mapl-rollout-end}
        \State Repeat lines~\ref{line:mapl-rollout-start}--%
        \ref{line:mapl-rollout-end} for the same task using
        $\pi_{\theta_i}$ if $\nu=0$, otherwise $\varpi_i$,
        to obtain $(\mathbf{h}'_H,\tau'_H)$.
        \State Annotate the pair; discard ties and add
        non-tied pairs to $\mathcal{D}^{(\nu)}$.
    \EndFor
    \State $\mathcal{B}^{(\nu)} \gets$ top-$k$ pairs
    in $\mathcal{D}^{(\nu)}$.
        \State Initialize reward model $\widetilde{R}_{\psi}$
        if $\nu=0$.
        \For{each reward-model training step}
            \State Sample $b$ preference pairs
            from $\mathcal{B}^{(\nu)}$.
            \State Compute per-pair losses
            $\mathcal{L}_{\mathrm{RM}}^{(j)}(\psi)$
            according to Equation~\ref{eq:marlhf-rm},
            $j=1,\ldots,b$.
            \State $\psi \gets \psi
            - \alpha_R \frac{1}{b}
            \sum_{j=1}^{b}
            \nabla_{\psi}\mathcal{L}_{\mathrm{RM}}^{(j)}(\psi)$.
        \EndFor
        \State Hold the reward model $\widetilde{R}_{\psi}$ and reference policy $\boldsymbol{\pi}_{\mathrm{ref}}$ fixed.
        \For{each policy training step}
            \State $\boldsymbol{\pi}_{\mathrm{old}} \gets \boldsymbol{\pi}_{\boldsymbol{\theta}}$.
            \State Sample $b$ tasks from $\mathcal{X}$.
            \For{each sampled task}
                \State Collect $G$ independent joint histories
                under $\boldsymbol{\pi}_{\mathrm{old}}$ by repeating
                lines~\ref{line:mapl-rollout-start}--%
                \ref{line:mapl-rollout-end}.
                \State Compute each trajectory's predicted return
                $\sum_{t=0}^{H-1}
                \widetilde{R}_{\psi}(\mathbf{h}_t^{(g)},\mathbf{a}_t^{(g)})$,
                $g=1,\ldots,G$.
                \State Compute group-relative advantages by subtracting
                the group mean predicted return.
            \EndFor
        \BeginBox[draw=gray, line width=0.6pt]
            \State Compute per-task losses
            $\mathcal{L}_{\mathrm{MARL}^{\mathcal{G}}}^{(j)}
            (\boldsymbol{\theta})$
            according to Equation~\ref{eq:marlhf},
            $j=1,\ldots,b$.
            \For{each agent $i\in\mathcal{I}$ in parallel}
                \State $\theta_i \gets \theta_i
                - \alpha_\pi \frac{1}{b}
                \sum_{j=1}^{b}
                \nabla_{\theta_i}
                \mathcal{L}_{\mathrm{MARL}^{\mathcal{G}}}^{(j)}
                (\boldsymbol{\theta})$.
            \EndFor
        \EndBox
    \EndFor
\EndFor
\State \Return $\{\pi_{\theta_i}\}_{i\in\mathcal{I}}$.
\end{algorithmic}
\end{algorithm}

\begin{algorithm}[t]
\caption{MADPO under Dec-Dec}
\label{alg:madpo}
\small
\begin{algorithmic}[1]
\State \textbf{Input:} Task dataset $\mathcal{X}$;
LLM agents $\{\pi_{\theta_i}\}_{i\in\mathcal{I}}$;
LLM comparators $\{\varpi_i\}_{i\in\mathcal{I}}$;
LLM reference policy $\boldsymbol{\pi}_{\mathrm{ref}}$;
human annotator following oracle reward $R^*$;
learning rate $\alpha_\pi$;
training batch size $b$;
pair count $m$; buffer size $k$;
horizon $H$; iterations $\mathcal{T}$; KL coefficient $\eta>0$.
\For{iteration $\nu = 0,\ldots,\mathcal{T}-1$}
    \State $\mathcal{D}^{(\nu)} \gets \emptyset$.
    \For{each task $x\sim\mathcal{X}$
    until $|\mathcal{D}^{(\nu)}|=m$}
        \State Collect $(\mathbf{h}_H,\tau_H)$ following
        lines~\ref{line:mapl-rollout-start}--%
        \ref{line:mapl-rollout-end}
        of Algorithm~\ref{alg:mapl}.
        \State Repeat these lines for the same task using
        $\pi_{\theta_i}$ if $\nu=0$, otherwise $\varpi_i$,
        to obtain $(\mathbf{h}'_H,\tau'_H)$.
        \State Annotate the pair; discard ties and add
        non-tied pairs to $\mathcal{D}^{(\nu)}$.
    \EndFor
    \State $\mathcal{B}^{(\nu)} \gets$ top-$k$ pairs
    in $\mathcal{D}^{(\nu)}$.
    \State Hold the reference policy $\boldsymbol{\pi}_{\mathrm{ref}}$ fixed.
    \For{each training step}
        \State Sample $b$ preference pairs
        from $\mathcal{B}^{(\nu)}$.
        \BeginBox[draw=gray, line width=0.6pt]
            \State Compute per-pair losses
            $\mathcal{L}_{\mathrm{MADPO}}^{(j)}
            (\boldsymbol{\theta})$
            according to Equation~\ref{eq:madpo},
            $j=1,\ldots,b$.
            \For{each agent $i\in\mathcal{I}$ in parallel}
                \State $\theta_i \gets \theta_i
                - \alpha_\pi \frac{1}{b}
                \sum_{j=1}^{b}
                \nabla_{\theta_i}
                \mathcal{L}_{\mathrm{MADPO}}^{(j)}
                (\boldsymbol{\theta})$.
            \EndFor
        \EndBox
    \EndFor
\EndFor
\State \Return $\{\pi_{\theta_i}\}_{i\in\mathcal{I}}$.
\end{algorithmic}
\end{algorithm}

\subsection{Derivation of MARLHF}
\label{app:marlhf}

Considering a fixed history-based reward $\widetilde{R}_{\psi}(\mathbf{h},\mathbf{a})$, we maximize the expected cumulative reward with KL-regularization,
\begin{equation}
\mathcal{J}(\boldsymbol{\theta})
=
\mathbb{E}_{\mathbf{h}_H\sim\boldsymbol{\pi}_{\boldsymbol{\theta}}}
\left[
\sum_{t=0}^{H-1}
\widetilde{R}_{\psi}(\mathbf{h}_t,\mathbf{a}_t)
-\eta\log
\frac{
\boldsymbol{\pi}_{\boldsymbol{\theta}}(\mathbf{h}_H)
}{
\boldsymbol{\pi}_{\mathrm{ref}}(\mathbf{h}_H)
}
\right].
\nonumber
\end{equation}
Since
$\mathbb{E}_{\mathbf{h}_H\sim\boldsymbol{\pi}_{\boldsymbol{\theta}}}
[\nabla_{\boldsymbol{\theta}}
\log\boldsymbol{\pi}_{\boldsymbol{\theta}}(\mathbf{h}_H)]\!=\!0$, differentiating the sampling distribution and the log-ratio gives
\begin{equation}
\nabla_{\boldsymbol{\theta}}\mathcal{J}(\boldsymbol{\theta})
=
\mathbb{E}_{\mathbf{h}_H\sim\boldsymbol{\pi}_{\boldsymbol{\theta}}}
\left[
\left(
\sum_{t=0}^{H-1}\widetilde{R}_{\psi}(\mathbf{h}_t,\mathbf{a}_t)
-\eta\log
\frac{\boldsymbol{\pi}_{\boldsymbol{\theta}}(\mathbf{h}_H)}
{\boldsymbol{\pi}_{\mathrm{ref}}(\mathbf{h}_H)}
\right)
\nabla_{\boldsymbol{\theta}}
\log\boldsymbol{\pi}_{\boldsymbol{\theta}}(\mathbf{h}_H)
\right].
\nonumber
\end{equation}
We independently sample $G>1$ histories for the same task under $\boldsymbol{\pi}_{\mathrm{old}}\gets\boldsymbol{\pi}_{\boldsymbol{\theta}_{\mathrm{old}}}$. Subtracting the group mean gives the group-relative advantage,
\begin{equation}
\widetilde{A}^{(g)}
=
\sum_{t=0}^{H-1}
\left[
\widetilde{R}_{\psi}(\mathbf{h}_t^{(g)},\mathbf{a}_t^{(g)})
-\frac{1}{G}\sum_{g'=1}^{G}
\widetilde{R}_{\psi}(\mathbf{h}_t^{(g')},\mathbf{a}_t^{(g')})
\right].
\nonumber
\end{equation}

We obtain the KL-gradient term by treating the log-ratio as a stop-gradient weight:
\begin{equation}
\nabla_{\boldsymbol{\theta}}
\left[
\eta\mathsf{sg}\!\left[
\log\frac{
\boldsymbol{\pi}_{\boldsymbol{\theta}}(\mathbf{h}_H)
}{
\boldsymbol{\pi}_{\mathrm{ref}}(\mathbf{h}_H)
}
\right]
\log\boldsymbol{\pi}_{\boldsymbol{\theta}}(\mathbf{h}_H)
\right]
=
\eta\log\frac{
\boldsymbol{\pi}_{\boldsymbol{\theta}}(\mathbf{h}_H)
}{
\boldsymbol{\pi}_{\mathrm{ref}}(\mathbf{h}_H)
}
\nabla_{\boldsymbol{\theta}}
\log\boldsymbol{\pi}_{\boldsymbol{\theta}}(\mathbf{h}_H).
\nonumber
\end{equation}
Combining the corrected reward advantage and KL weight
gives the surrogate in Equation~\ref{eq:marlhf}:
\begin{equation}
\mathcal{L}_{\mathrm{MARL}^{\mathcal{G}}}(\boldsymbol{\theta})
=
-\mathbb{E}_{\mathbf{h}_H^{\mathcal{G}}\sim\boldsymbol{\pi}_{\mathrm{old}}}
\left[
\frac{1}{G}\sum_{g=1}^{G}
\mathsf{sg}\!\left[
\frac{G}{G-1}\widetilde{A}^{(g)}
-\eta\log
\frac{
\boldsymbol{\pi}_{\boldsymbol{\theta}}(\mathbf{h}_H^{(g)})
}{
\boldsymbol{\pi}_{\mathrm{ref}}(\mathbf{h}_H^{(g)})
}
\right]
\log\boldsymbol{\pi}_{\boldsymbol{\theta}}(\mathbf{h}_H^{(g)})
\right].
\nonumber
\end{equation}

In decentralized MAS, agents have separate parameters and independently generate responses from their local histories. Differentiating the joint loss with respect to agent $i$'s parameters gives
\begin{equation}
\nabla_{\theta_i}
\mathcal{L}_{\mathrm{MARL}^{\mathcal{G}}}(\boldsymbol{\theta})
=
-\mathbb{E}_{\mathbf{h}_H^{\mathcal{G}}\sim\boldsymbol{\pi}_{\mathrm{old}}}
\left[
\frac{1}{G}\sum_{g=1}^{G}
\left(
\frac{G}{G-1}\widetilde{A}^{(g)}
-\eta\log
\frac{\boldsymbol{\pi}_{\boldsymbol{\theta}}(\mathbf{h}_H^{(g)})}
{\boldsymbol{\pi}_{\mathrm{ref}}(\mathbf{h}_H^{(g)})}
\right)
\sum_{t=0}^{H-1}
\nabla_{\theta_i}
\log\pi_{\theta_i}(a_{i,t}^{(g)}|h_{i,t}^{(g)})
\right].
\nonumber
\end{equation}

In centralized MAS, the coordinator generates joint responses from joint histories. Differentiating the loss with respect to its parameters $\boldsymbol{\theta}$ gives
\begin{equation}
\nabla_{\boldsymbol{\theta}}
\mathcal{L}_{\mathrm{MARL}^{\mathcal{G}}}(\boldsymbol{\theta})
=
-\mathbb{E}_{\mathbf{h}_H^{\mathcal{G}}\sim\boldsymbol{\pi}_{\mathrm{old}}}
\left[
\frac{1}{G}\sum_{g=1}^{G}
\left(
\frac{G}{G-1}\widetilde{A}^{(g)}
-\eta\log
\frac{\boldsymbol{\pi}_{\boldsymbol{\theta}}(\mathbf{h}_H^{(g)})}
{\boldsymbol{\pi}_{\mathrm{ref}}(\mathbf{h}_H^{(g)})}
\right)
\sum_{t=0}^{H-1}
\nabla_{\boldsymbol{\theta}}
\log\boldsymbol{\pi}_{\boldsymbol{\theta}}
(\mathbf{a}_t^{(g)}|\mathbf{h}_t^{(g)})
\right].
\nonumber
\end{equation}

\subsection{Derivation of MADPO}
\label{app:madpo-derivation}

We optimize the same objective as MARLHF above. Since the current and reference policies share the initial distribution and environment dynamics, the expected sum of log-ratios is the KL divergence between their full history distributions.

We assume arbitrary distributions over full histories can be realized by the reference class, then
\begin{equation}
\begin{aligned}
\prod_{t=0}^{H-1}
\frac{
\boldsymbol{\pi}^{*}(\mathbf{a}_t|\mathbf{h}_t)
}{
\boldsymbol{\pi}_{\mathrm{ref}}(\mathbf{a}_t|\mathbf{h}_t)
}
&=
\frac{1}{Z}
\exp\left(
\frac{1}{\eta}
\sum_{t=0}^{H-1}
\widetilde{R}_{\psi}(\mathbf{h}_t,\mathbf{a}_t)
\right),
\\
\sum_{t=0}^{H-1}
\widetilde{R}_{\psi}(\mathbf{h}_t,\mathbf{a}_t)
&=
\eta\sum_{t=0}^{H-1}
\log
\frac{
\boldsymbol{\pi}^{*}(\mathbf{a}_t|\mathbf{h}_t)
}{
\boldsymbol{\pi}_{\mathrm{ref}}(\mathbf{a}_t|\mathbf{h}_t)
}
+\eta\log Z,
\end{aligned}
\nonumber
\end{equation}
where $Z=\mathbb{E}_{\mathbf{h}_H\sim\boldsymbol{\pi}_{\mathrm{ref}}} [\exp(\frac{1}{\eta}\sum_{t=0}^{H-1} \widetilde{R}_{\psi}(\mathbf{h}_t,\mathbf{a}_t))]$ normalizes the full history distribution. 

For two histories for the same task, the $\eta\log Z$ terms cancel by subtracting their returns. A policy-based implicit reward can then be constructed,
\begin{equation}
\widetilde{R}_{\boldsymbol{\theta}}(\mathbf{h},\mathbf{a})
=
\eta\log
\frac{
\boldsymbol{\pi}_{\boldsymbol{\theta}}(\mathbf{a}|\mathbf{h})
}{
\boldsymbol{\pi}_{\mathrm{ref}}(\mathbf{a}|\mathbf{h})
},\,
\sum_{t=0}^{H-1}
\widetilde{R}_{\boldsymbol{\theta}}(\mathbf{h}_t,\mathbf{a}_t)
=
\eta\sum_{t=0}^{H-1}
\log
\frac{
\boldsymbol{\pi}_{\boldsymbol{\theta}}(\mathbf{a}_t|\mathbf{h}_t)
}{
\boldsymbol{\pi}_{\mathrm{ref}}(\mathbf{a}_t|\mathbf{h}_t)
}.
\nonumber
\end{equation}

We define the preference margin
$\Delta_{\boldsymbol{\theta}}
=\frac{1}{\eta}\sum_{t=0}^{H-1}
[\widetilde{R}_{\boldsymbol{\theta}}(\mathbf{h}_t^w,\mathbf{a}_t^w)
-\widetilde{R}_{\boldsymbol{\theta}}(\mathbf{h}_t^l,\mathbf{a}_t^l)]$,
so that $\eta\Delta_{\boldsymbol{\theta}}$ is the difference
between the two cumulative implicit rewards.
Applying the BT model and averaging over preference pairs
in $\mathcal{B}^{(\nu)}$ gives Equation~\ref{eq:madpo},
\begin{equation}
\mathcal{L}_{\mathrm{MADPO}}(\boldsymbol{\theta})
=
-\mathbb{E}_{\mathcal{B}^{(\nu)}}
\left[
\log\sigma\left(
\eta\sum_{t=0}^{H-1}
\left(
\log
\frac{
\boldsymbol{\pi}_{\boldsymbol{\theta}}(\mathbf{a}_t^w|\mathbf{h}_t^w)
}{
\boldsymbol{\pi}_{\mathrm{ref}}(\mathbf{a}_t^w|\mathbf{h}_t^w)
}
-
\log
\frac{
\boldsymbol{\pi}_{\boldsymbol{\theta}}(\mathbf{a}_t^l|\mathbf{h}_t^l)
}{
\boldsymbol{\pi}_{\mathrm{ref}}(\mathbf{a}_t^l|\mathbf{h}_t^l)
}
\right)
\right)
\right].
\nonumber
\end{equation}

In decentralized MAS, each agent $\pi_{\theta_i}$ generates its
response from its local history, so the margin of contribution is,
\begin{equation}
\Delta_{\theta_i}
=
\sum_{t=0}^{H-1}
\left(
\log
\frac{
\pi_{\theta_i}(a_{i,t}^w|h_{i,t}^w)
}{
\pi_{\mathrm{ref},i}(a_{i,t}^w|h_{i,t}^w)
}
-
\log
\frac{
\pi_{\theta_i}(a_{i,t}^l|h_{i,t}^l)
}{
\pi_{\mathrm{ref},i}(a_{i,t}^l|h_{i,t}^l)
}
\right).
\nonumber
\end{equation}
Since $\frac{d}{dz}\log\sigma(z)=1-\sigma(z)=\sigma(-z)$,
the chain rule gives the MADPO gradient for each agent:
\begin{equation}
\begin{aligned}
\nabla_{\theta_i}
\mathcal{L}_{\mathrm{MADPO}}(\boldsymbol{\theta})
&=
-\eta\mathbb{E}_{\mathcal{B}^{(\nu)}}
\left[
\left(1-\sigma(\eta\Delta_{\boldsymbol{\theta}})\right)
\nabla_{\theta_i}\Delta_{\boldsymbol{\theta}}
\right]
\\
&=
-\eta\mathbb{E}_{\mathcal{B}^{(\nu)}}
\left[
\sigma(-\eta\Delta_{\boldsymbol{\theta}})
\nabla_{\theta_i}\Delta_{\theta_i}
\right].
\end{aligned}
\nonumber
\end{equation}

In centralized MAS, a coordinator generates the
joint response from the joint history.
We apply the same chain rule, but differentiate with respect
to the coordinator's parameters $\boldsymbol{\theta}$,
\begin{equation}
\nabla_{\boldsymbol{\theta}}
\mathcal{L}_{\mathrm{MADPO}}(\boldsymbol{\theta})
=
-\eta\mathbb{E}_{\mathcal{B}^{(\nu)}}
\left[
\sigma(-\eta\Delta_{\boldsymbol{\theta}})
\sum_{t=0}^{H-1}
\nabla_{\boldsymbol{\theta}}\log
\frac{
\boldsymbol{\pi}_{\boldsymbol{\theta}}(\mathbf{a}_t^w|\mathbf{h}_t^w)
}{
\boldsymbol{\pi}_{\boldsymbol{\theta}}(\mathbf{a}_t^l|\mathbf{h}_t^l)
}
\right].
\nonumber
\end{equation}

%% file: appendix/variants.tex
\section{MAPL Variants} \label{app:variants}


In \S~\ref{sec:mapl_methods}, we select the top-$k$ preference pairs ranked by their winning trajectories' returns under $R^*$. Here, we describe alternatives from two aspects: which comparator generates the candidate trajectories and which samples are retained for training.

\subsection{Comparator Policy}

At iteration $\nu$, MAPL compares the agent policy $\boldsymbol{\pi}^{(\nu)}$ with a comparator policy $\boldsymbol{\varpi}^{(\nu)}$. The comparator may belong to the same policy class $\boldsymbol{\Pi}$ as the agents. Examples include an earlier checkpoint, $\boldsymbol{\varpi}^{(\nu)}=\boldsymbol{\pi}^{(\nu')}$ for $0\leqslant\nu'<\nu$, or a higher-temperature variant of the current policy, $\boldsymbol{\varpi}^{(\nu)}=\boldsymbol{\pi}^{(\nu)}_{\mathrm{soft}}$. Alternatively, the comparator may belong to another policy class $\boldsymbol{\mathcal{U}}$, using a different model architecture or from different information structures, e.g., Dec-Cen.

\subsection{Sampling Strategy}

\begin{wrapfigure}{r}{0.5\linewidth}
    \centering
    \captionsetup{font=footnotesize}
    \vspace{-5mm}
    \includegraphics[width=\linewidth]{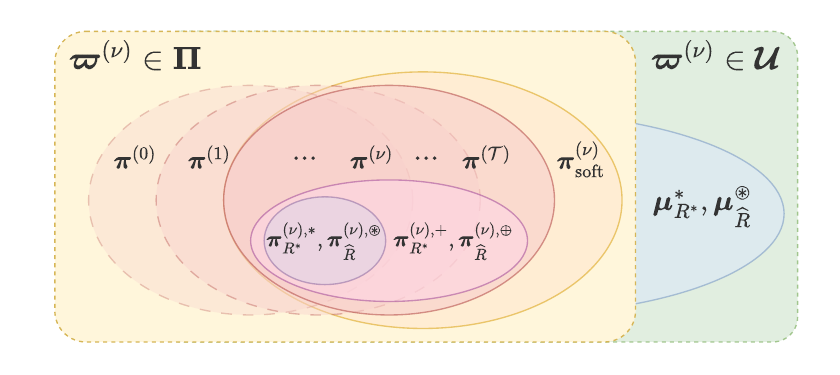}
    \vspace{-10mm}
    \caption{A Venn diagram of different comparator policies.\vspace{-2mm}}
    \label{fig:venn}
\end{wrapfigure}
Rejection sampling is commonly used in LLM post-training to curate training data for RLHF~\citep{nakano2021webgpt,dong2023raft}. In \S~\ref{sec:mapl_methods}, we select the top samples according to $R^*$ and the retained pairs $\mathcal{B}^{(\nu)}\!\sim\!\{\boldsymbol{\pi}_{R^*}^{(\nu),*}\!\lor\!\boldsymbol{\varpi}_{R^*}^{(\nu),*}\}$.
In practice, however, $R^*$ may not be easily accessible when the criteria underlying annotator preferences are unknown. Selection can instead rely on a proxy $\widetilde{R}$, such as a learned reward model or a human-specified metric. In this case, the replay buffer can be selected by $\mathcal{B}^{(\nu)}\!\sim\!\{\boldsymbol{\pi}_{\widetilde{R}}^{(\nu),\circledast}\lor\boldsymbol{\varpi}_{\widetilde{R}}^{(\nu),\circledast}\}$, where circled asterisks denote the best performance under a proxy reward model $\widetilde{R}$. Although including preferred trajectories from the agents' own policies is beneficial for reinforcing a good policy through exploitation, we often expect the agent to explore beneficial features beyond their current policy class by learning from the comparator's trajectories at later stages. Another form of rejection sampling includes ``good'' comparator samples that meet at least one such criterion, denoted as $\mathcal{B}^{(\nu)}\sim\boldsymbol{\varpi}_{R^*}^{(\nu),+}$ or $\mathcal{B}^{(\nu)}\sim\boldsymbol{\varpi}_{\widetilde{R}}^{(\nu),\oplus}$, where $+$ and $\oplus$ denote that the comparator's trajectory is at least as good in one aspect under $R^*$ or $\widetilde{R}$.

%% file: appendix/exp-setting.tex
\section{Experimental Settings}
\label{app:exp_setting}

\subsection{Task and Model Configurations}

Table~\ref{tab:model_config} summarizes the tasks and model configurations in Table~\ref{tab:main_results},\ref{tab:main-ablation} and Figure~\ref{fig:main-results},\ref{fig:main-ablation}. For \texttt{BFCL}, we retain samples where multi-agent cooperation is meaningful. For \texttt{Travel}, we use short trips as the performance differences among optimization methods can be distinguished with relatively small agents. This also excludes samples with excessively long contexts to reduce VRAM requirements. All methods are evaluated on the same filtered subsets.

\begin{table}[ht]
\centering
\begin{minipage}{0.98\textwidth}
\centering
\captionsetup{font=footnotesize}
\setlength{\tabcolsep}{4pt}
\renewcommand{\arraystretch}{1}
\resizebox{\linewidth}{!}{%
\begin{tabular}{lcccc}
\toprule
\textbf{Configuration} & \texttt{TLDR} & \texttt{CoopHE} & \texttt{BFCL} & \texttt{Travel} \\
\midrule
\rowcolor{pptint}
\multicolumn{5}{l}{\textit{\textbf{Task}}} \\
Dataset & \textit{trl-lib}\textit{/tldr} & \textit{OpenMLRL/CoopHumanEval} & \textit{BFCL} & \textit{osunlp/TravelPlanner}\\
Filter & None & None & \textit{parallel\;\textup{\&}\;parallel\_multiple} & \textit{days$\leqslant$5\;\textup{\&}\;visiting\_city\_num$\leqslant$2} \\
Training & \textit{train}[0,250) & \textit{train}[16:82) & \textit{data}[0:160) & \textit{validation}[0:60) \\
Evaluation & \textit{eval}[0, 250) & \textit{train}[0:16) & \textit{data}[160:200) & \textit{validation}[60:82) \\

\midrule
\rowcolor{pptint}
\multicolumn{5}{l}{\textbf{Model}} \\
Agent (Dec) & \textit{Qwen3-1.7B} & \textit{Qwen2.5-Coder-3B} & \textit{Qwen3-4B-Instruct-2507} & \textit{Qwen3-4B-Instruct-2507} \\
Agent (Cen) & \textit{Qwen3-4B} & \textit{Qwen2.5-Coder-7B} & \textit{Qwen3-8B} & \textit{Qwen3-8B} \\
Temperature & 0.7 & 0.6 & 0.7 & 0.7 \\
Top-$k$ & 20 & 50 & 20 & 20 \\
Top-$p$ & 0.9 & 0.6 & 0.8 & 0.8 \\
Max output tokens & 256 & 256 & 256 & 1024 \\
Torch data type & bfloat16 & bfloat16 & bfloat16 & bfloat16 \\
Attention backend & sdpa & sdpa & eager & sdpa \\
\bottomrule
\end{tabular}%
}
\vspace{-2mm}
\caption{Task and model configurations. Unless otherwise specified, the agents, comparators, and reward model in MARLHF share the same model configuration. The filtered subsets preserve the original order.}
\label{tab:model_config}
\end{minipage}
\end{table}

\subsection{Hyperparameters}

Table~\ref{tab:hyper} shows the hyperparameters used in Table~\ref{tab:main_results} and Figure~\ref{fig:main-results}. For ablation studies, we change the preference dataset pairs and decay factor $\lambda$ to obtain the results in Table~\ref{tab:main-ablation}(a), the comparator model in Table~\ref{tab:main-ablation}(b), and MARL algorithms in Table~\ref{tab:main-ablation}(c).

\begin{table}[ht]
\centering
\begin{minipage}{0.8\textwidth}
\centering
\captionsetup{font=footnotesize}
\setlength{\tabcolsep}{5pt}
\renewcommand{\arraystretch}{1}
\resizebox{0.98\linewidth}{!}{%
\begin{tabular}{lcccc}
\toprule
\textbf{Hyperparameter} & \texttt{TLDR} & \texttt{CoopHE} & \texttt{BFCL} & \texttt{Travel} \\
\midrule

\rowcolor{pptint}
\multicolumn{5}{l}{\textit{\textbf{Comparator}}} \\
Generation mode (Dec) & decentralized & decentralized & decentralized & decentralized \\
Generation mode (Cen) & centralized & centralized & centralized & centralized \\
Model & \textit{current} & \textit{current} & \textit{current\_copy} & \textit{current\_copy} \\
Preference dataset pairs & 20 & 20 & 20 & 20 \\

\midrule
\rowcolor{pptint}
\multicolumn{5}{l}{\textit{\textbf{Replay Buffer}}} \\
Replay buffer pairs & 4 & 4 & 4 & 4 \\
Decay factor $\lambda$ (MARLHF) & 0.8 & 0.8 & 0.8 & 0.8 \\
Decay factor $\lambda$ (MADPO) & 0.2 & 0.2 & 0.2 & 0.2 \\

\midrule
\rowcolor{pptint}
\multicolumn{5}{l}{\textbf{Agent Optimization}} \\
Optimizer & \textit{AdamW} & \textit{AdamW} & \textit{AdamW} & \textit{AdamW} \\
Policy learning rate & $8\!\times\!10^{-6}$ & $2\!\times\!10^{-5}$ & $2\!\times\!10^{-5}$ & $2\!\times\!10^{-5}$ \\
Policy training batch size & 4 & 4 & 4 & 8 \\
Policy training epochs (MARLHF) & 2 & 4 & 2 & 3 \\
Policy training epochs (MADPO) & 1 & 2 & 1 & 1 \\
MARL algorithm (MARLHF) & MAGRPO & MAGRPO & MAGRPO & MAGRPO  \\
MAGRPO group size $G$ (MARLHF) & 4 & 4 & 4 & 4 \\
Advantage normalization & True & True & True & False \\
KL penalty $\eta$ & 0.1 & 0.1 & 0.1 & 0.1 \\

\midrule
\rowcolor{pptint}
\multicolumn{5}{l}{\textbf{Reward Model (MARLHF)}} \\
Reward learning rate & $10^{-5}$ & $10^{-5}$ & $10^{-5}$ & $10^{-5}$ \\
Reward training batch size & 4 & 4 & 1 & 1 \\
Reward training epochs & 4 & 2 & 1 & 1 \\
Max input tokens & 1024 & 2048 & 4096 & 32768 \\
\bottomrule
\end{tabular}%
}
\vspace{-2mm}
\caption{Comparator, replay buffer, and optimization hyperparameters.}
\label{tab:hyper}
\end{minipage}
\end{table}

\subsection{Oracle Rewards} \label{app:dataset_details}

Following previous studies \citep{christiano2017deep,lee2021pebble,kim2023preference,zhu2024decoding}, we assume annotators label their preferences according to the following oracle rewards.

\paragraph{TLDR}
The oracle reward of \texttt{TLDR} considers structural quality, style consistency, and logical coherence. Both summaries must contain at least 8 tokens to receive an initial reward of 0.5 and proceed to
the following evaluation. Then we check the length ratio and the unique-word ratio of the second agent's response to the first's. Since the second agent is complementary, a length ratio between 1.6 and 3.2 and a unique-word ratio above 2.0 obtain full rewards of 1.5, while ratios falling within the tolerances of 1.1 to 5.0 and 1.3 to 2.0 are scaled down proportionally. Responses outside this scope receive no rewards. We further measure style consistency by the Jaccard similarity of vocabularies between the two responses, with a score of 0.6. Coherence is measured by the occurrence of transition words, with a score of 0.4. The total reward of \texttt{TLDR} ranges from 0 to 3.

\paragraph{CoopHE}For \texttt{CoopHE}, we evaluate the structural validity, test pass rate, and the call rate. A base reward of 0.5 is given once agents output valid code snippets. Then we check the validity of the auxiliary and main function signatures, where successful ones receive 0.4 and 0.6 rewards, respectively. The concatenated code snippets are parsed using \textit{ast} to check for syntax errors. Syntactically correct code receives a reward of 1 and is then run against the unit tests under a 5-second execution limit. The runtime correctness reward is scaled proportionally to the pass rate. Finally, we evaluate cooperation quality, giving a bonus of 1.5 when the main function actually calls the auxiliary function without merely acting as a wrapper. The reward scale of \texttt{CoopHE} is from 0 to 4.

\paragraph{BFCL}
We consider call validity, correctness, and coordination in \texttt{BFCL} reward. A base reward of 0.1 is given when at least one tool call can be extracted. Validity and correctness are scored with a weight of 1.5 by comparing the predicted function names with the reference signatures and checking the predicted arguments. Coordination is scored with weight 0.4, based on whether the agents issue exactly the expected number of calls and whether their contributions (\#calls) are balanced. To avoid reward hacking, we further refine the reward by penalizing duplicated calls, inactive agents, and excess calls with weights of -0.6, -0.3, and -0.1, respectively. \texttt{BFCL} rewards scale from -1 to 2.

\paragraph{Travel}
Each agent is assigned either the logistics role or the activity role and must use the provided reference information for its assigned role. For valid individual and joint outputs, it could receive up to 0.06 in total. Specifically, we evaluate both agents' contributions with a weight of 0.14, coverage of role-specific required fields with a weight of 0.08, and the quality of the itinerary with a weight of 0.57. The itinerary is examined on completeness, reference grounding, consistency across routes and cities, diversity of dining and attractions, minimum required stays, and alignment with budget and user preferences. A bonus reward of 0.10 is given when both agents fulfill all requirements. Partial rewards are also given for partial format compliance and recoverable assignments, with up to 0.05. Penalties are applied for invalid assignments, conflicting or duplicated entries, entirely copying the references, and overly long outputs, with weights of 0.10, 0.05, 0.05, 0.10, and 0.05. Finally, the total reward is clipped to [-0.25, 1].

%% file: appendix/additional.tex
\section{Additional Results}
\label{app:additional_results}

As discussed in \S\ref{sec:experiments_ablation}, MARLHF and MADPO exhibit different sensitivities to hyperparameters. In this section, we analyze several key hyperparameters in detail.

\subsection{Ablation on Dataset Sizes} \label{subapp:dataset_size}

\begin{figure}[ht]
    \centering
    \captionsetup{font=footnotesize}
    \includegraphics[width=0.88\textwidth]{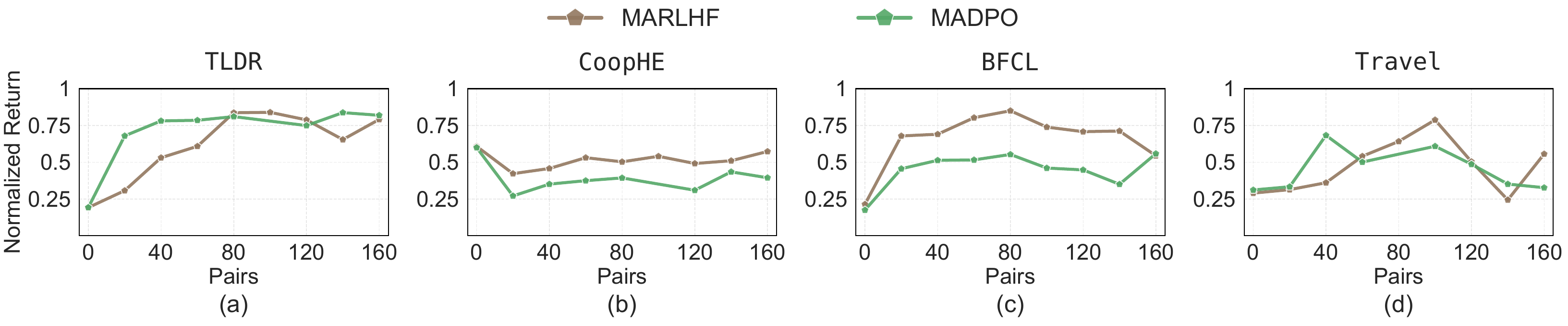}
    \vspace{-3mm}
    \caption{Ablation on different sizes of $\mathcal{D}$ for offline MARLHF and MADPO (\textcolor{brown}{brown} and \textcolor{green}{green}) across 5 runs.}
    \label{fig:ablation_dataset_size}
\end{figure}

We first conduct an ablation on the size of the offline preference dataset for MAPL methods. As shown in Figure~\ref{fig:ablation_dataset_size}, providing more data generally improves MARLHF and MADPO on \texttt{TLDR} and \texttt{BFCL} as expected. This is more pronounced for MARLHF, since building a complete reward representation requires substantially more data. More data also helps for MADPO, as it likely provides higher-quality data that MADPO can imitate. 

Notably, this trend does not hold for \texttt{CoopHE} and \texttt{Travel}. On \texttt{CoopHE}, no matter how many preference pairs are added, training still gets stuck at a local optimum in which one agent performs all the tasks while the other remains idle. We attribute this to the raw model, \textit{Qwen2.5-Coder}, which does not inherently cooperate. So without iterative training, it rarely provides samples with effective cooperative schemes. Both MADPO and MARLHF initially benefit from additional samples on \texttt{Travel} (roughly from 40 to 100 samples), but this improvement drops off beyond 120 samples. In this task, increasing the sample size might dilute the proportion of the few high-quality samples originally in the preference dataset. Therefore, it could even hurt training performance. 

In summary, increasing the number of samples does not necessarily help MAPL training. It increases the computational overhead and can dilute the proportion of valuable samples in the dataset. In practice, the appropriate dataset size thus depends on both the domain and the raw model.

\subsection{Ablation on Replay Buffers} \label{subapp:ablation_buffer}

We show the details of ablation on replay buffers in Table~\ref{tab:main-ablation}(a). We consider two types of replay buffers. The first constructs the replay buffer from the preference dataset collected $k$ iterations earlier. The second samples across all iterations throughout training, using a decay factor $\lambda$ to downweight older data. Setting $\lambda=1$ yields uniform sampling across these datasets, while $k=0$ in the first scheme and $\lambda=0$ in the second both restrict sampling to the current iteration.

\begin{figure}[ht]
    \centering
    \captionsetup{font=footnotesize}
    \includegraphics[width=0.88\textwidth]{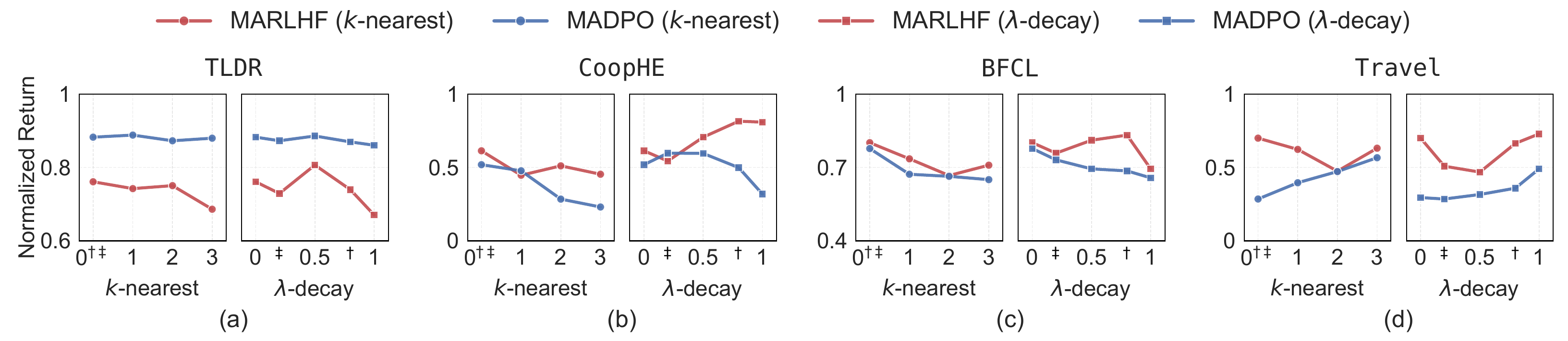}
    \vspace{-3mm}
    \caption{Ablation on the different replay buffers for MARLHF (\textcolor{red}{red}) and MADPO (\textcolor{blue}{blue}) across 5 runs. We consider the decentralized MAS. $^\dagger$ and $^\ddagger$ denote the pivot settings for MARLHF and MADPO, respectively.}
    \label{fig:ablation_replay_buffer_all}
\end{figure}

\begin{figure}[ht]
    \centering
    \captionsetup{font=footnotesize}
    \includegraphics[width=0.88\textwidth]{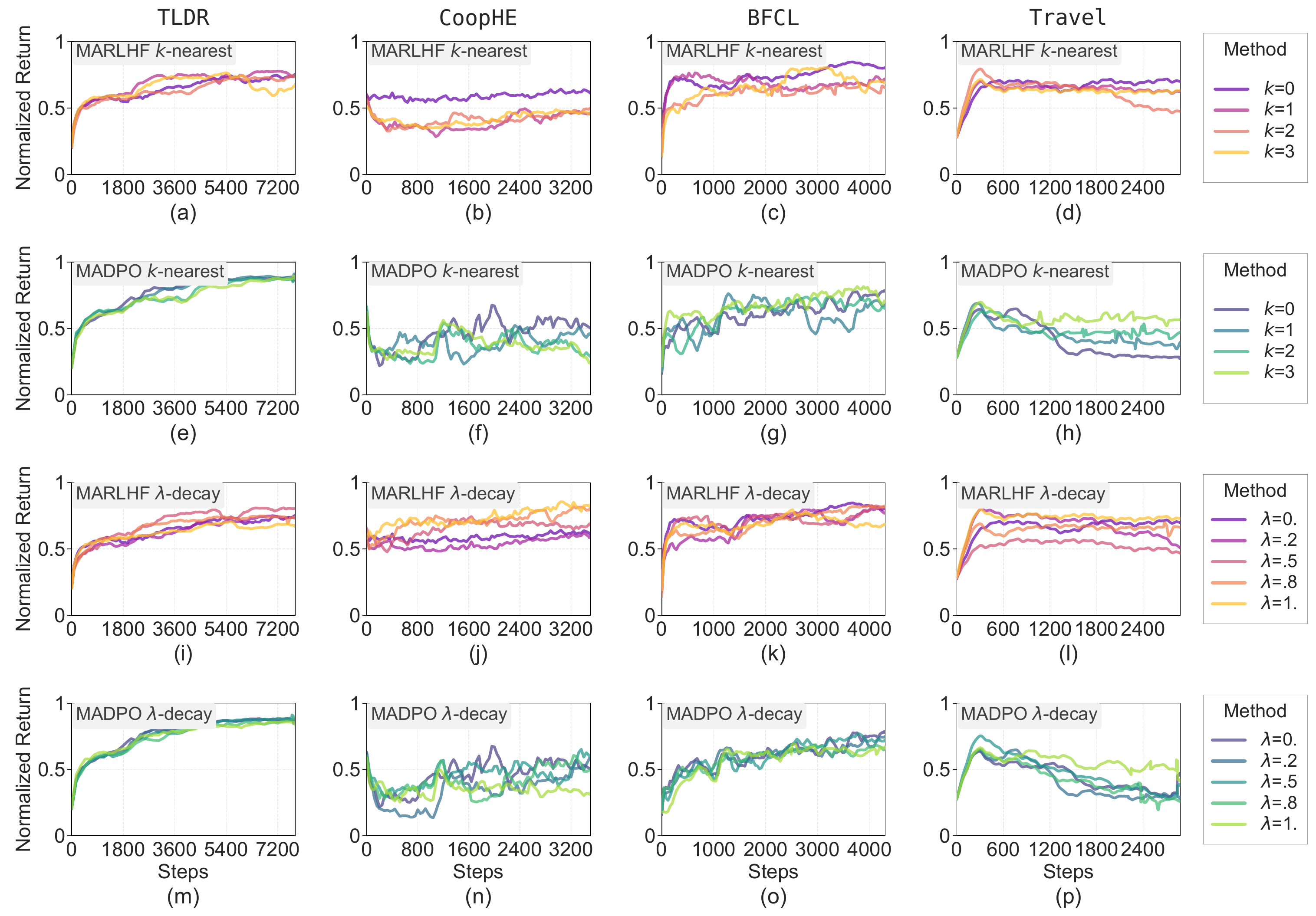}
    \caption{Results of MARLHF and MADPO on \texttt{TLDR}, \texttt{CoopHE}, \texttt{BFCL}, and \texttt{Travel} across 5 runs.}
    \vspace{-3mm}
    \label{fig:ablation_buffer}
\end{figure}

Figure~\ref{fig:ablation_replay_buffer_all} shows that increasing $k$ (i.e., using older data) decreases the performance of both MARLHF and MADPO on \texttt{CoopHE} and \texttt{BFCL}. This suggests that a more recently fine-tuned model that generates higher-quality data improves performance for both MAPL algorithms. This pattern is less clear on \texttt{TLDR}, where both curves are flat with $k$. Since training converges quickly in this domain, even earlier datasets already contain a high density of good samples, so performance across different $k$ is hard to distinguish. On \texttt{Travel}, the pattern reverses for MADPO. This is because MADPO training collapses in the middle stage, so earlier datasets may contain a higher concentration of high-quality samples, which is opposite to the trend observed in other domains.

Incorporating richer samples from different training stages generally improves training, as they provide agents with more solution features, enabling more effective collaboration. Although the best $\lambda$ varies across domains, it is consistently higher for MARLHF than for MADPO. Across all four domains, the best $\lambda$ for MARLHF exceeds 0.5, whereas it ranges from 0 to 0.2 for MADPO in most tasks, which implies that MARLHF may require more diverse data to construct a reward model with a complete representation, whereas MADPO benefits more from a higher density of high-quality samples in the dataset (\S~\ref{sec:experiments}). We present the results during training for Figure~\ref{fig:ablation_replay_buffer_all} in Figure~\ref{fig:ablation_buffer}.

\subsection{Ablation on Comparators} \label{subapp:ablation_comparator}

\begin{figure}[ht]
    \centering
    \captionsetup{font=footnotesize}
    \includegraphics[width=0.88\textwidth]{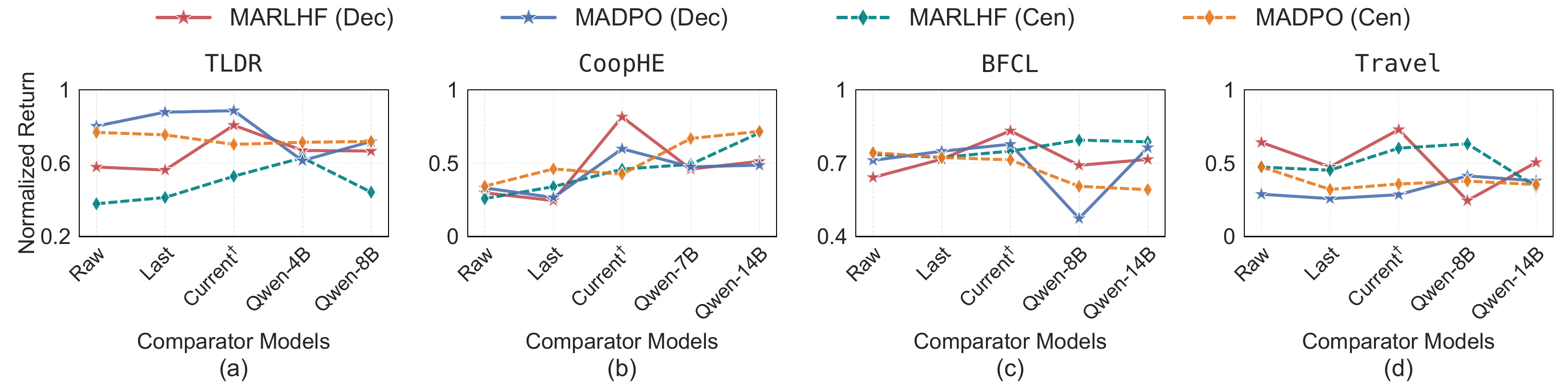}
    \vspace{-3mm}
    \caption{Ablation on the different comparators used in MARLHF and MADPO across 5 runs, $\boldsymbol{\varpi}_\mathrm{dec}$ in \textcolor{red}{red} and \textcolor{blue}{blue}, $\boldsymbol{\varpi}_\mathrm{cen}$ in \textcolor{bluegreen}{blue-green} and \textcolor{orange}{orange}, respectively. We consider decentralized MAS. $^\dagger$ denotes the pivot settings.}
    \label{fig:ablation_comparator_all}
\end{figure}

We show the details of ablation on comparators presented in Table~\ref{tab:main-ablation}(b). Figure~\ref{fig:ablation_comparator_all} shows results using different models as comparators, as well as a comparison between centralized and decentralized comparators. Along the axes, \textit{Raw} denotes the raw agent model, \textit{Last} the agent checkpoint from the previous iteration, and \textit{Current} the current agent model. The first three columns show the MAPL performance by using an increasingly updated model as the comparator. Across \texttt{TLDR}, \texttt{CoopHE}, and \texttt{BFCL}, all MAPL algorithms benefit from later comparator checkpoints under a decentralized comparator, as reflected by the upward trend of the solid curves from left to right. This is because later models provide higher-quality samples that benefit training. We find that this does not consistently hold on \texttt{Travel}, where solid lines are flat or oscillating. MADPO tends to collapse in later training stages (as shown in the last 1500 steps in Figure~\ref{fig:main-ablation}) and therefore can no longer produce higher-quality samples. This improvement also does not extend to the centralized comparator, as the dashed lines are flat along these three columns on \texttt{CoopHE} and \texttt{BFCL}.

\begin{figure}[ht]
    \centering
    \captionsetup{font=footnotesize}
    \includegraphics[width=0.88\textwidth]{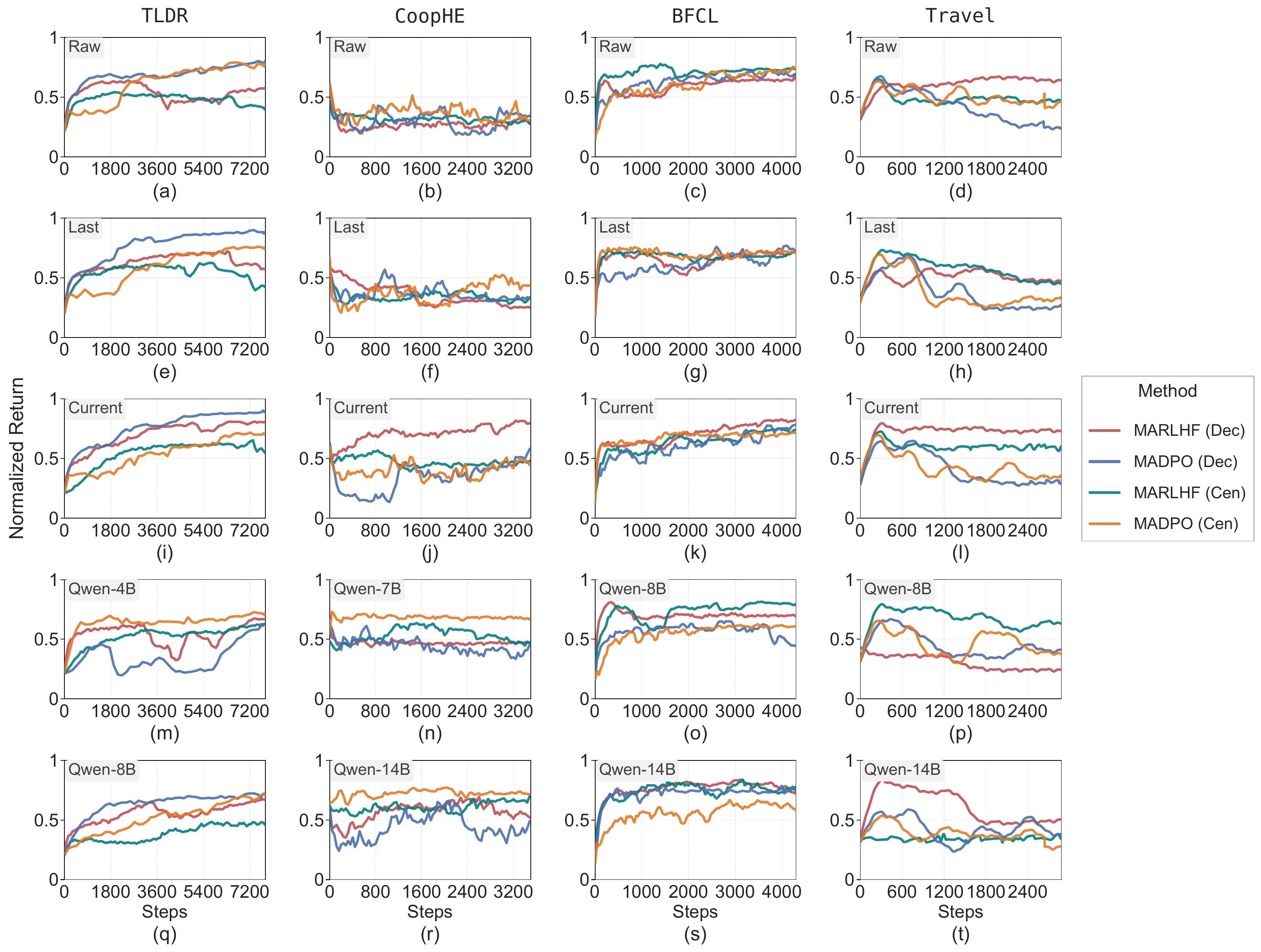}
    \caption{Results of MARLHF and MADPO on TLDR, CoopHE, BFCL, and Travel across 5 runs.}
    \vspace{-3mm}
    \label{fig:ablation_comparator}
\end{figure}

We find that most of the solid curves drop from the third column to the fourth. This suggests that using a larger model under the same policy class may not even provide as much improvement to training as using the currently trained model as a comparator. This is because the samples provided by the larger model, while higher quality, may require capabilities that the smaller model simply cannot achieve. This also echoes the second challenge in \S~\ref{sec:background}. Even if an existing fixed offline multi-agent preference dataset could be acquired, its utility for training is likely limited. We do not observe a consistent pattern as to whether a larger model helps more as a decentralized or as a centralized comparator across these domains. We present the results along the training for Figure~\ref{fig:ablation_comparator_all} in Figure~\ref{fig:ablation_comparator}.

\subsection{Ablation on MAS Configurations}

The ablations in \S\ref{sec:experiments_ablation} consider homogeneous MAS, where all agents share the same model. We extend the analysis to MAS composed of agents with different model sizes. We consider two settings: heterogeneous MAS, where only one agent (either the first or the second) has twice the parameter count of its counterpart, and homogeneous-scaled MAS, in which both agents' sizes are increased.

\begin{figure}[ht]
    \centering
    \captionsetup{font=footnotesize}
    \includegraphics[width=0.88\textwidth]{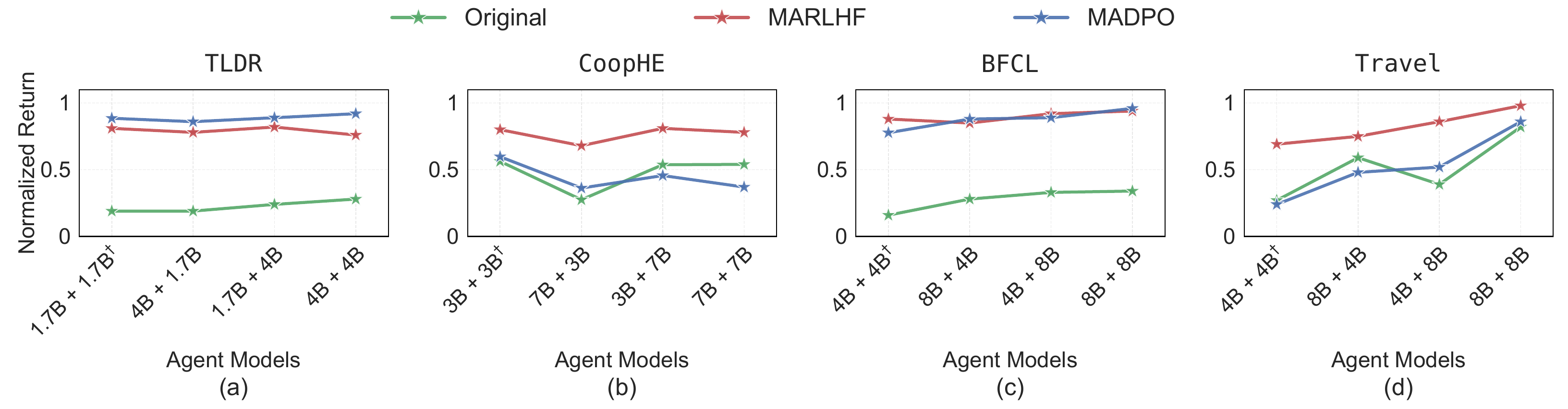}
    \vspace{-3mm}
    \caption{Ablation on different MAS configurations. \textcolor{green}{Green} curves present the performance before training, \textcolor{red}{red} and \textcolor{blue}{blue} show results via MARLHF and MADPO training. Hyperparameters are the same as in Table~\ref{tab:hyper}.}
    \label{fig:ablation_mas}
\end{figure}

Figure~\ref{fig:ablation_mas} reports the performance of these configurations before training and after MAPL training. On \texttt{TLDR}, larger agents achieve higher performance before training, but the gain from MAPL training is marginal as model size increases. Given that both MAPL algorithms already perform well in MAS with smaller models, there is little room for improvement. On \texttt{CoopHE}, increasing agent size even degrades it in the ablated configurations. We attribute this to the limited capability of the base models in this task, where the pretrained LLMs likely included few collaborative samples regardless of scale. Moreover, when using the same number of samples as in Table~\ref{tab:hyper}, it is much more difficult to optimize for larger LLMs. On \texttt{BFCL} and \texttt{Travel}, performance improves steadily with model size, both before and after MAPL training, which demonstrates the effectiveness of MAPL across a broader range of MAS.

\subsection{Results on Larger-Scale MAS}

We further ablate MAPL on larger-scale MAS. Due to the dual-role nature of \texttt{TLDR} and \texttt{CoopHE}, where the first agent mostly
serves as an auxiliary agent and the second generates the main solution, adding more agents into these MAS might significantly change the task prompt and reward design. We conduct this ablation on \texttt{BFCL} and \texttt{Travel} for a fair comparison.

\begin{wraptable}{r}{0.54\textwidth}
\centering
\setlength{\tabcolsep}{4pt}
\renewcommand{\arraystretch}{1.2}
\resizebox{0.98\linewidth}{!}{
\vspace{-2mm}
\begin{tabular}{l@{\hspace{10pt}}cccc}
\toprule
\multirow{2}{*}[-0.5ex]{\textbf{Method}}
& \multicolumn{2}{c}{\textbf{BFCL}}
& \multicolumn{2}{c}{\textbf{Travel}} \\
\cmidrule(lr){2-3}
\cmidrule(lr){4-5}
& Two-Agent$^\dagger$
& Three-Agent
& Two-Agent$^\dagger$
& Three-Agent \\
\midrule

Raw MAS
& 12.8
& 17.5
& 66.2
& 63.1 \\
MARLHF
& \textbf{22.7}
& \textbf{18.4}
& \textbf{70.5}
& \textbf{72.6} \\
MADPO
& 15.6
& 15.1
& 41.5
& 44.3 \\

\bottomrule
\end{tabular}
}
\captionsetup{font=footnotesize}
\vspace{-2mm}
\caption{Performance of two-agent and three-agent MAS on BFCL and Travel. We report Correct for BFCL and Pass for Travel. Raw MAS denotes parallel inference with raw LLMs.
}
\label{tab:agent-count-ablation}
\end{wraptable}
For \texttt{BFCL}, the task setup remains largely unchanged, except that three agents now jointly make the function calls. Introducing an additional agent yields a slight improvement in raw MAS. But this gain is not preserved after MAPL optimization. Both MARLHF and MADPO underperform the two-agent MAS. This is because training a reward model over the longer context induced by three agents becomes more difficult with the same amount of preferences. Also, credit assignment becomes more complex for larger-scale MAS optimization. For \texttt{Travel}, we split the single activity agent into two agents, one responsible for dining and the other for sightseeing, while keeping the output format and evaluation protocol unchanged. Adding this agent does not improve the performance of the raw MAS. After MAPL optimization, however, both MARLHF and MADPO allow the three-agent MAS to surpass the two-agent MAS. This shows that MAPL can effectively optimize larger-scale MAS, though the extent of improvement is highly task-dependent.

%% file: appendix/prompt.tex
\section{Prompt Design} \label{app:prompt}

We present the prompts used in Table~\ref{tab:main_results} for \texttt{BFCL} and \texttt{Travel} as follows. The prompt template for \texttt{TLDR} and \texttt{CoopHE} follows \citet{liu2026llm}.

\begin{promptbox}{\texttt{BFCL Prompt Design - Agent 0}}
You are a decentralized function-calling agent.

2 agents are answering independently. Hence, you cannot communicate with the other agent, and you cannot see their answer. Later, a downstream aggregator will merge the agents' tool calls.

<<Your assignment:>>
Another agent is working on the same request in parallel, so you do not need to handle every part of the task yourself. Focus on the parts you can address most confidently and make any tool calls that are useful for those parts. Aim to contribute meaningful progress rather than covering everything, and avoid returning no contribution when there is something you can do.

<<BFCL native category:>> (*@\promptarg{CATEGORY}@*)
<<Heuristic task type:>> (*@\promptarg{TASK_TYPE}@*)

<<Available function schemas:>>
(*@\promptarg{FUNCTION_SCHEMAS}@*)

<<User request:>>
(*@\promptarg{USER_REQUEST}@*)

<<Output requirements:>>
- Output only the tool calls assigned to you or selected by you under your assignment.
- Use Python-style function-call syntax, one call per line.
- Use keyword arguments from the function schemas.
- Do not include explanations, markdown, numbering, or reasoning.
- If you make no tool call, output exactly: []

<<Example output format:>>
function_name(arg1="value", arg2=3)
another.function_name(flag=True)
\end{promptbox}

\begin{promptbox}{\texttt{BFCL Prompt Design - Agent 1}}
You are a decentralized function-calling agent.

2 agents are answering independently. Hence, you cannot communicate with the other agent, and you cannot see their answer. Later, a downstream aggregator will merge the agents' tool calls.

<<Your assignment:>>
Another agent is working on the same request in parallel, so you do not need to handle every part of the task yourself. Focus on the parts you can address most confidently and make any tool calls that are useful for those parts. Aim to contribute meaningful progress rather than covering everything, and avoid returning no contribution when there is something you can do.

<<BFCL native category:>> (*@\promptarg{CATEGORY}@*)
<<Heuristic task type:>> (*@\promptarg{TASK_TYPE}@*)

<<Available function schemas:>>
(*@\promptarg{FUNCTION_SCHEMAS}@*)

<<User request:>>
(*@\promptarg{USER_REQUEST}@*)

<<Output requirements:>>
- Output only the tool calls assigned to you or selected by you under your assignment.
- Use Python-style function-call syntax, one call per line.
- Use keyword arguments from the function schemas.
- Do not include explanations, markdown, numbering, or reasoning.
- If you make no tool call, output exactly: []

<<Example output format:>>
function_name(arg1="value", arg2=3)
another.function_name(flag=True)
\end{promptbox}

\begin{promptbox}{\texttt{Travel Prompt Design - Agent 0}}
[System message]
You are one agent in a decentralized travel-planning team. Follow the user's role and output contract exactly. Return only the requested JSON object. Close the assignments array with ] before the final top-level.

[User message]
You are Agent 0 in a decentralized travel-planning team.

There are 2 agents in total, and every agent receives the same trip request and acts together. You cannot communicate with the other agent or see its output. A deterministic downstream merger combines all slot assignments into one itinerary. Every agent receives the same reward for that merged itinerary.

<<ROLE GUIDANCE:>>
You are in charge of <<LOGISTICS AND FEASIBILITY>> only: current_city, transportation, and accommodation for every day. Fill every owned slot, including an explicit '-' when the itinerary convention permits it to be empty. Do not fill breakfast, attraction, lunch, or dinner slots owned by Agent 1.

<<YOUR OWNED SLOTS>>:
(*@\promptarg{OWNED_SLOTS}@*)
(*@\promptarg{ROLE_SLOT_CHECKLIST}@*)
(*@\promptarg{ROLE_HARD_CONSTRAINT_DUTY}@*)
(*@\promptarg{BUDGET_CONTRACT}@*)

<<COLLABORATION RULES:>>
- The itinerary has (*@\promptarg{NUM_TOTAL_SLOTS}@*) slots: (*@\promptarg{DAYS}@*) days x 7 fields.
- You may submit at most (*@\promptarg{NUM_OWNED_SLOTS}@*) assignments.
- Submit exactly (*@\promptarg{NUM_OWNED_SLOTS}@*) assignments: one for every slot assigned to your role.
- A slot is identified by (day, field) and should be assigned exactly once by the team.
- Same-slot duplication wastes capacity; different values for one slot create a conflict.
- If a slot needs to be intentionally empty, explicitly assign the string "-". An omitted slot is considered missing, which is different from explicitly assigning "-".
- Write current_city as "from A to B" on a travel day and as the single current city on a stay day. The trip starts at the origin, visits the requested number of cities, and returns to the origin on the final day.
- A travel day requires matching transportation; a stay day can use "-" for transportation.
- A stay day requires breakfast, attraction, lunch, and dinner. These experience fields may be "-" on a travel day. Accommodation is required on every day except the final return day.
- Never use "-" merely because you are uncertain. Use it only in the cases above.
- Use only entities and facts from the <<TRIP REQUEST>>, <<STRUCTURED CONSTRAINTS>>, and <<REFERENCE-DERIVED PLANNING CONTEXT>>.
- Do not make up prices, flight numbers, restaurants, attractions, or accommodations.
- Write restaurants, attractions, and accommodations as "Name, City". Never combine multiple attractions: choose exactly one grounded attraction on a required stay day. Copy a complete transportation candidate, including its mode or flight number and the matching route, rather than returning an abbreviation.
- Your own section is not scored separately; optimize the final team itinerary.

<<VALID FIELDS:>>
current_city, transportation, breakfast, attraction, lunch, dinner, accommodation

<<TRIP REQUEST>>:
(*@\promptarg{TRIP_REQUEST}@*)

<<STRUCTURED CONSTRAINTS>>:
origin=(*@\promptarg{ORIGIN}@*)
destination=(*@\promptarg{DESTINATION}@*)
days=(*@\promptarg{DAYS}@*)
dates=(*@\promptarg{DATES}@*)
people=(*@\promptarg{NUM_PEOPLE}@*)
budget=(*@\promptarg{BUDGET}@*)
local_constraints=(*@\promptarg{LOCAL_CONSTRAINTS}@*)

<<REFERENCE-DERIVED PLANNING CONTEXT>>:
(*@\promptarg{REFERENCE_CONTEXT}@*)

<<STRICT OUTPUT CONTRACT:>>
- Your entire response must be exactly one JSON object: no Markdown fence, prefix, explanation, note, suffix, or second object.
- The object must have exactly two keys: "agent_id" and "assignments".
- "agent_id" must be the integer 0.
- "assignments" must contain exactly (*@\promptarg{NUM_OWNED_SLOTS}@*) objects. It must never exceed (*@\promptarg{NUM_OWNED_SLOTS}@*) objects.
- Every assignment object must have exactly three keys: "day" (integer), "field" (one valid field), and "value" (string copied from the reference, or "-").
- Do not copy schema wording or placeholder text as a value.
- The generation system has already written the <<ASSISTANT PREFILL>> below. It is part of your response, so do not repeat any portion of it.
- The prefill ends immediately after the opening quote of your first owned value. Generate only that value's contents, then close the value with a double quote.
- After each closing value quote, the generation system supplies the next fixed assignment object in <<YOUR OWNED SLOTS>> order. It fixes every day, field, key, comma, bracket, and brace; you choose only each value.
- Do not try to generate the next assignment's schema yourself. Immediately continue inside the next already-open value string. The system closes the JSON object after the final owned value.
- A per-value token limit prevents one slot from consuming the whole response budget. Keep every value concise and copy catalog spelling exactly.
- The reconstructed response always has the fixed schema and assignment count shown above. Semantic omissions and invalid values still receive low reward.

<<ASSISTANT PREFILL>> (already supplied; do not repeat it):
{"agent_id": 0, "assignments": [{"day": 1, "field": "current_city", "value": "

Continue the prefilled JSON now. Your first generated character is the first character inside the already-open value string.
\end{promptbox}

\begin{promptbox}{\texttt{Travel Prompt Design - Agent 1}}
[System message]
You are one agent in a decentralized travel-planning team. Follow the user's role and output contract exactly. Return only the requested JSON object. Close the assignments array with ] before the final top-level.

[User message]
You are Agent 1 in a decentralized travel-planning team.

There are 2 agents in total, and every agent receives the same trip request and acts together. You cannot communicate with the other agent or see its output. A deterministic downstream merger combines all slot assignments into one itinerary. Every agent receives the same reward for that merged itinerary.

<<ROLE GUIDANCE:>>
You are in charge of <<DAILY EXPERIENCE>> only: breakfast, attraction, lunch, and dinner for every day. Fill every owned slot, including an explicit '-' when the itinerary convention permits it to be empty. Do not fill current_city, transportation, or accommodation slots owned by Agent 0.

<<YOUR OWNED SLOTS>>:
(*@\promptarg{OWNED_SLOTS}@*)
(*@\promptarg{ROLE_SLOT_CHECKLIST}@*)
(*@\promptarg{ROLE_HARD_CONSTRAINT_DUTY}@*)
(*@\promptarg{BUDGET_CONTRACT}@*)

<<COLLABORATION RULES:>>
- The itinerary has (*@\promptarg{NUM_TOTAL_SLOTS}@*) slots: (*@\promptarg{DAYS}@*) days x 7 fields.
- You may submit at most (*@\promptarg{NUM_OWNED_SLOTS}@*) assignments.
- Submit exactly (*@\promptarg{NUM_OWNED_SLOTS}@*) assignments: one for every slot assigned to your role.
- A slot is identified by (day, field) and should be assigned exactly once by the team.
- Same-slot duplication wastes capacity; different values for one slot create a conflict.
- If a slot needs to be intentionally empty, explicitly assign the string "-". An omitted slot is considered missing, which is different from explicitly assigning "-".
- Write current_city as "from A to B" on a travel day and as the single current city on a stay day. The trip starts at the origin, visits the requested number of cities, and returns to the origin on the final day.
- A travel day requires matching transportation; a stay day can use "-" for transportation.
- A stay day requires breakfast, attraction, lunch, and dinner. These experience fields may be "-" on a travel day. Accommodation is required on every day except the final return day.
- Never use "-" merely because you are uncertain. Use it only in the cases above.
- Use only entities and facts from the <<TRIP REQUEST>>, <<STRUCTURED CONSTRAINTS>>, and <<REFERENCE-DERIVED PLANNING CONTEXT>>.
- Do not make up prices, flight numbers, restaurants, attractions, or accommodations.
- Write restaurants, attractions, and accommodations as "Name, City". Never combine multiple attractions: choose exactly one grounded attraction on a required stay day. Copy a complete transportation candidate, including its mode or flight number and the matching route, rather than returning an abbreviation.
- Your own section is not scored separately; optimize the final team itinerary.

<<VALID FIELDS:>>
current_city, transportation, breakfast, attraction, lunch, dinner, accommodation

<<TRIP REQUEST>>:
(*@\promptarg{TRIP_REQUEST}@*)

<<STRUCTURED CONSTRAINTS>>:
origin=(*@\promptarg{ORIGIN}@*)
destination=(*@\promptarg{DESTINATION}@*)
days=(*@\promptarg{DAYS}@*)
dates=(*@\promptarg{DATES}@*)
people=(*@\promptarg{NUM_PEOPLE}@*)
budget=(*@\promptarg{BUDGET}@*)
local_constraints=(*@\promptarg{LOCAL_CONSTRAINTS}@*)

<<REFERENCE-DERIVED PLANNING CONTEXT>>:
(*@\promptarg{REFERENCE_CONTEXT}@*)

<<STRICT OUTPUT CONTRACT:>>
- Your entire response must be exactly one JSON object: no Markdown fence, prefix, explanation, note, suffix, or second object.
- The object must have exactly two keys: "agent_id" and "assignments".
- "agent_id" must be the integer 1.
- "assignments" must contain exactly (*@\promptarg{NUM_OWNED_SLOTS}@*) objects. It must never exceed (*@\promptarg{NUM_OWNED_SLOTS}@*) objects.
- Every assignment object must have exactly three keys: "day" (integer), "field" (one valid field), and "value" (string copied from the reference, or "-").
- Do not copy schema wording or placeholder text as a value.
- The generation system has already written the <<ASSISTANT PREFILL>> below. It is part of your response, so do not repeat any portion of it.
- The prefill ends immediately after the opening quote of your first owned value. Generate only that value's contents, then close the value with a double quote.
- After each closing value quote, the generation system supplies the next fixed assignment object in <<YOUR OWNED SLOTS>> order. It fixes every day, field, key, comma, bracket, and brace; you choose only each value.
- Do not try to generate the next assignment's schema yourself. Immediately continue inside the next already-open value string. The system closes the JSON object after the final owned value.
- A per-value token limit prevents one slot from consuming the whole response budget. Keep every value concise and copy catalog spelling exactly.
- The reconstructed response always has the fixed schema and assignment count shown above. Semantic omissions and invalid values still receive low reward.

<<ASSISTANT PREFILL>> (already supplied; do not repeat it):
{"agent_id": 1, "assignments": [{"day": 1, "field": "breakfast", "value": "

Continue the prefilled JSON now. Your first generated character is the first character inside the already-open value string.
\end{promptbox}

%% file: appendix/compute.tex
\clearpage
\section{Compute Resources} \label{app:resources}

We summarize the hardware configurations and estimated training costs of representative experiments in Table~\ref{tab:training_cost}. Decentralized agents are easier to deploy and train on separate devices, whereas centralized training requires substantial VRAM without sharding, since the centralized policy must process all agents' combined information and is typically larger than any individual decentralized agent. This suggests that decentralized collaboration can be a practical solution when tasks are modularly separable and with compute is limited.

\begin{table}[h]
\centering
\begin{minipage}{0.89\textwidth}
\centering
\captionsetup{font=footnotesize}
\setlength{\tabcolsep}{6pt}
\renewcommand{\arraystretch}{1}
\setlength{\dashlinedash}{2pt}
\setlength{\dashlinegap}{2pt}
\resizebox{\linewidth}{!}{%
\begin{tabular}{lccccccccc}
\toprule
\multirow{2}{*}{\textbf{Method}} &
\multirow{2}{*}{GPUs} &
\multirow{2}{*}{Duration} &
\multicolumn{3}{c}{Model Placement} &
\multicolumn{4}{c}{Peak GPU VRAM} \\
\cmidrule(lr){4-6}\cmidrule(lr){7-10}
& & & Agents & Reward & Comparator & cuda:0 & cuda:1 & cuda:2 & Total \\

\midrule
\rowcolor{pptint}
\multicolumn{10}{l}{\textit{\textbf{TLDR}}} \\
MAGRPO (Dec) & 1$\times$RTX & 12 h & cuda:0 & \textit{N/A} & \textit{N/A} & 40 & \textit{N/A} & \textit{N/A} & 40 \\
MAGRPO (Cen) & 1$\times$B200 & 10 h & cuda:0 & \textit{N/A} & \textit{N/A} & 50 & \textit{N/A} & \textit{N/A} & 50 \\
\hdashline
MARLHF (Offline) & 1$\times$B200 & 12 h & cuda:0 & cuda:0 & \textit{N/A} & 55 & \textit{N/A} & \textit{N/A} & 55 \\
MARLHF (Dec-Dec) & 1$\times$RTX & 15 h & cuda:0 & cuda:0 & cuda:0 & 75 & \textit{N/A} & \textit{N/A} & 75 \\
MARLHF (Dec-Cen) & 1$\times$RTX & 15 h & cuda:0 & cuda:0 & cuda:0 & 75 & \textit{N/A} & \textit{N/A} & 75 \\
MARLHF (Cen-Cen) & 1$\times$B200 & 10 h & cuda:0 & cuda:0 & cuda:0 & 105 & \textit{N/A} & \textit{N/A} & 105 \\
\hdashline
MADPO (Offline) & 1$\times$RTX & 5 h & cuda:0 & \textit{N/A} & \textit{N/A} & 50 & \textit{N/A} & \textit{N/A} & 50 \\
MADPO (Dec-Dec) & 1$\times$RTX & 7 h & cuda:0 & \textit{N/A} & cuda:0 & 50 & \textit{N/A} & \textit{N/A} & 50 \\
MADPO (Dec-Cen) & 1$\times$RTX & 7 h & cuda:0 & \textit{N/A} & cuda:0 & 50 & \textit{N/A} & \textit{N/A} & 50 \\
MADPO (Cen-Cen) & 1$\times$B200 & 7 h & cuda:0 & \textit{N/A} & cuda:0 & 60 & \textit{N/A} & \textit{N/A} & 60 \\

\midrule
\rowcolor{pptint}
\multicolumn{10}{l}{\textit{\textbf{CoopHE}}} \\
MAGRPO (Dec) & 1$\times$H200 & 5 h & cuda:0 & \textit{N/A} & \textit{N/A} & 70 & \textit{N/A} & \textit{N/A} & 70 \\
MAGRPO (Cen) & 1$\times$B200 & 2 h & cuda:0 & \textit{N/A} & \textit{N/A} & 70 & \textit{N/A} & \textit{N/A} & 70 \\
\hdashline
MARLHF (Offline) & 1$\times$H200 & 2 h & cuda:0 & cuda:0 & \textit{N/A} & 135 & \textit{N/A} & \textit{N/A} & 135 \\
MARLHF (Dec-Dec) & 1$\times$B200 & 3 h & cuda:0 & cuda:0 & cuda:0 & 180 & \textit{N/A} & \textit{N/A} & 180 \\
MARLHF (Dec-Cen) & 1$\times$B200 & 2 h & cuda:0 & cuda:0 & cuda:0 & 180 & \textit{N/A} & \textit{N/A} & 180 \\
MARLHF (Cen-Cen) & 1$\times$B200 & 2 h & cuda:0 & cuda:0 & cuda:0 & 180 & \textit{N/A} & \textit{N/A} & 180 \\
\hdashline
MADPO (Offline) & 1$\times$B200 & 1 h & cuda:0 & \textit{N/A} & \textit{N/A} & 70 & \textit{N/A} & \textit{N/A} & 70 \\
MADPO (Dec-Dec) & 1$\times$B200 & 2 h & cuda:0 & \textit{N/A} & cuda:0 & 75 & \textit{N/A} & \textit{N/A} & 75 \\
MADPO (Dec-Cen) & 1$\times$B200 & 1 h & cuda:0 & \textit{N/A} & cuda:0 & 75 & \textit{N/A} & \textit{N/A} & 75 \\
MADPO (Cen-Cen) & 1$\times$B200 & 1 h & cuda:0 & \textit{N/A} & cuda:0 & 100 & \textit{N/A} & \textit{N/A} & 100 \\

\midrule
\rowcolor{pptint}
\multicolumn{10}{l}{\textit{\textbf{BFCL}}} \\
MAGRPO (Dec) & 2$\times$B200 & 4 h & cuda:0,1 & \textit{N/A} & \textit{N/A} & 150 & 120 & \textit{N/A} & 270 \\
MAGRPO (Cen) & 2$\times$B200 & 7 h & cuda:0,1 & \textit{N/A} & \textit{N/A} & 135 & 140 & \textit{N/A} & 275 \\
\hdashline
MARLHF (Offline) & 3$\times$B200 & 4 h & cuda:0,1 & cuda:2 & \textit{N/A} & 115 & 110 & 180 & 405 \\
MARLHF (Dec-Dec) & 3$\times$B200 & 10 h & cuda:0,1 & cuda:2 & cuda:2 & 180 & 180 & 180 & 540 \\
MARLHF (Dec-Cen) & 3$\times$B200 & 8 h & cuda:0,1 & cuda:2 & cuda:2 & 165 & 140 & 180 & 485 \\
MARLHF (Cen-Cen) & 3$\times$B200 & 12 h & cuda:0,1 & cuda:2 & cuda:2 & 140 & 155 & 140 & 435 \\
\hdashline
MADPO (Offline) & 2$\times$B200 & 5 h & cuda:0,1 & \textit{N/A} & \textit{N/A} & 140 & 140 & \textit{N/A} & 280 \\
MADPO (Dec-Dec) & 3$\times$B200 & 9 h & cuda:0,1 & \textit{N/A} & cuda:2 & 180 & 180 & 65 & 425 \\
MADPO (Dec-Cen) & 3$\times$B200 & 7 h & cuda:0,1 & \textit{N/A} & cuda:2 & 145 & 135 & 115 & 395 \\
MADPO (Cen-Cen) & 2$\times$B200 & 8 h & cuda:0,1 & \textit{N/A} & cuda:1 & 145 & 170 & \textit{N/A} & 315 \\

\midrule
\rowcolor{pptint}
\multicolumn{10}{l}{\textit{\textbf{Travel}}} \\
MAGRPO (Dec) & 1$\times$B200 & 4 h & cuda:0 & \textit{N/A} & \textit{N/A} & 90 & \textit{N/A} & \textit{N/A} & 90 \\
MAGRPO (Cen) & 1$\times$B200 & 5 h & cuda:0 & \textit{N/A} & \textit{N/A} & 110 & \textit{N/A} & \textit{N/A} & 110 \\
\hdashline
MARLHF (Offline) & 1$\times$RTX & 7 h & cuda:0 & cuda:0 & \textit{N/A} & 75 & \textit{N/A} & \textit{N/A} & 75 \\
MARLHF (Dec-Dec) & 2$\times$RTX & 15 h & cuda:0 & cuda:1 & cuda:1 & 100 & 60 & \textit{N/A} & 160 \\
MARLHF (Dec-Cen) & 1$\times$B200 & 9 h & cuda:0 & cuda:0 & cuda:0 & 100 & \textit{N/A} & \textit{N/A} & 100 \\
MARLHF (Cen-Cen) & 1$\times$B200 & 11 h & cuda:0 & cuda:0 & cuda:0 & 120 & \textit{N/A} & \textit{N/A} & 120 \\
\hdashline
MADPO (Offline) & 2$\times$H100 & 4 h & cuda:0,1 & \textit{N/A} & \textit{N/A} & 50 & 45 & \textit{N/A} & 95 \\
MADPO (Dec-Dec) & 1$\times$RTX & 11 h & cuda:0 & \textit{N/A} & cuda:0 & 85 & \textit{N/A} & \textit{N/A} & 85 \\
MADPO (Dec-Cen) & 1$\times$B200 & 5 h & cuda:0 & \textit{N/A} & cuda:0 & 95 & \textit{N/A} & \textit{N/A} & 95 \\
MADPO (Cen-Cen) & 1$\times$B200 & 4 h & cuda:0 & \textit{N/A} & cuda:0 & 100 & \textit{N/A} & \textit{N/A} & 100 \\
\bottomrule
\end{tabular}%
}
\vspace{-3mm}
\caption{Experimental devices and the peak GPU memory usage (GB) during training. RTX and H100 represent the RTX 6000 Pro Blackwell and H100 PCIe editions, respectively. The VRAM on each device is rounded to the nearest 5 GB, total VRAM sums the rounded values. }
\label{tab:training_cost}
\end{minipage}
\end{table}

%% file: appendix/related.tex
\section{Related Works}

\subsection{LLM-based MAS}

Many studies explore collaboration among specialized LLM agents to solve complex tasks. For example, agents act as programmers, reviewers, and testers for software development \citep{wu2023autogen,qian2024chatdev}. Other systems assign complementary responsibilities to query information \citep{chen2026improving}. In addition, LLM agents can also work together to conduct scientific research \citep{anthropic_multi_agent,schmidgall2025agent}. Most of these LLM-based MAS rely on centralized orchestration, with a predefined workflow coordinating agents. \citet{zhang2024proagent,chen2025internet,yang2026agentnet} study decentralized LLM-based MAS, where lightweight agents generate responses independently from their own local contexts. As they can be flexibly deployed on separate devices and run inference in parallel, it reduces sequential dependencies and hence improves efficiency. Meanwhile, the distributed deployment can also improve the robustness and security of the system \citep{liu2026learning}. We study preference-based optimization for both kinds of systems.

\subsection{Preference-based MARL}

Preference-based RL (PbRL) has been studied extensively to optimize policies from preference feedback~\citep{wirth2017survey}. Some work learns a reward function from preference signals and then optimizes the policy against it~\citep{zhu2024decoding,bui2025mapl,kang2025dual}, while other work instead trains a preference predictor that maps joint preferences to individual-level preferences used to optimize each agent's policy directly~\citep{kou2025offline}. However, assigning credit to individual agents and turns in MAS optimization is inherently difficult, and the resulting reward or value estimates are often inaccurate. Moreover, little of this work has been applied to LLM collaboration in practice, where the policy space is extremely large.

%% file: appendix/impact.tex
\section{Broader Impacts and Future Works}

Although MARL has recently been used to optimize LLM-based MAS, most existing work relies on manually curated reward models. Designing such models typically requires extensive domain knowledge, and their effectiveness in guiding training remains unclear. MAPL offers an alternative by learning from preference feedback, potentially reducing the need for task-specific reward design. This work opens the door to broadly optimizing multi-agent collaboration from human preference.

Nevertheless, our work has a few limitations for future study. As shown in Table~\ref{tab:training_cost}, MAPL requires substantially more compute than standard MARL fine-tuning, since it requires additional rollouts to adequately collect informative preferences for training. Though we propose an efficient instantiation MADPO,
it performs poorly on harder domains, such as \texttt{Travel}. Future work could address this gap or develop other efficient MAPL variants. Also, for larger-scale MAS and longer-horizon tasks, relying on preference feedback inevitably makes the training signal sparser. Yet, collecting a preference for every agent or every turn is impractical \citep{shani2405multi,kausik2025theoretical}. How to effectively scale MAPL to such settings remains unexplored. In addition, as discussed in \S\ref{sec:experiments}, the Dec-Cen paradigm introduces a realizability gap between the agents and the comparator. Developing methods that can facilitate transferring knowledge remains a direction for future work.